\documentclass[journal]{IEEEtran}
\usepackage{times}
\usepackage{bm}
\usepackage{soul}
\usepackage{url}
\usepackage{subfigure}
\usepackage{lineno,hyperref}
\usepackage[utf8]{inputenc}
\usepackage[small]{caption}
\usepackage{amsfonts}
\usepackage{graphicx}
\usepackage{amsmath}
\usepackage{multirow}
\usepackage{float}
\usepackage{amsthm}
\usepackage{booktabs}
\usepackage{algorithm}
\usepackage{xcolor}
\usepackage{algorithmic}
\usepackage{threeparttable}
\newtheorem{definition}{Definition}

\ifCLASSINFOpdf
\else
\fi
\begin{document}
\title{Hypernetwork Dismantling via Deep Reinforcement Learning}


\author{\IEEEauthorblockN{
Dengcheng Yan, Wenxin Xie, Yiwen Zhang, Qiang He, and Yun Yang
}
%\IEEEauthorblockA{
%\IEEEauthorrefmark{1} School of Computer Science and Technology, Anhui University, Hefei, China}

%\IEEEauthorrefmark{2} School of Software and Electrical Engineering, Swinburne University of Technology, Melbourne, Australia
% <-this % stops an unwanted space

\thanks{This work was supported by the National Natural Science Foundation of China (Grant No. U1936220, 61872002) and the University Natural Science Research Project of Anhui Province (Grant No. KJ2019A0037).}

\thanks{Dengcheng Yan, Wenxin Xie and Yiwen Zhang are with the School of Computer Science and Technology, Anhui University, Hefei 230039, China (e-mail: yanzhou@ahu.edu.cn, xiewxahu@foxmail.com, zhangyiwen@ahu.edu.cn). Corresponding author: Yiwen Zhang.}

\thanks{Qiang He and Yun Yang are with the School of Software and Electrical Engineering, Swinburne University of Technology, Melbourne 3122, Australia (e-mail:qhe@swin.edu.au, yyang@swin.edu.au).}
}

%{Shell \MakeLowercase{\textit{et al.}}: Bare Demo of IEEEtran.cls for IEEE %Transactions on Magnetics Journals}



% for Transactions on Magnetics papers, we must declare the abstract and
% index terms PRIOR to the title within the \IEEEtitleabstractindextext
% IEEEtran command as these need to go into the title area created by
\IEEEtitleabstractindextext{%
\begin{abstract}
Network dismantling aims to degrade the connectivity of a network by removing an optimal set of nodes. It has been widely adopted in many real-world applications such as epidemic control and rumor containment. However, conventional methods usually focus on simple network modeling with only pairwise interactions, while group-wise interactions modeled by hypernetwork are ubiquitous and critical. In this work, we formulate the hypernetwork dismantling problem as a node sequence decision problem and propose a deep reinforcement learning (DRL)-based hypernetwork dismantling framework. Besides, we design a novel inductive hypernetwork embedding method to ensure the transferability to various real-world hypernetworks. Our framework first generates small-scale synthetic hypernetworks and embeds the nodes and hypernetworks into a low dimensional vector space to represent the action and state space in DRL, respectively. Then trial-and-error dismantling tasks are conducted by an agent on these synthetic hypernetworks, and the dismantling strategy is continuously optimized. Finally, the well-optimized strategy is applied to real-world hypernetwork dismantling tasks. Experimental results on five real-world hypernetworks demonstrate the effectiveness of our proposed framework.
\end{abstract}

}

\maketitle

\IEEEdisplaynontitleabstractindextext

% \fi
\IEEEpeerreviewmaketitle

\section{Introduction}
\begin{figure}[htbp]
\centering
	\includegraphics[scale=0.3]{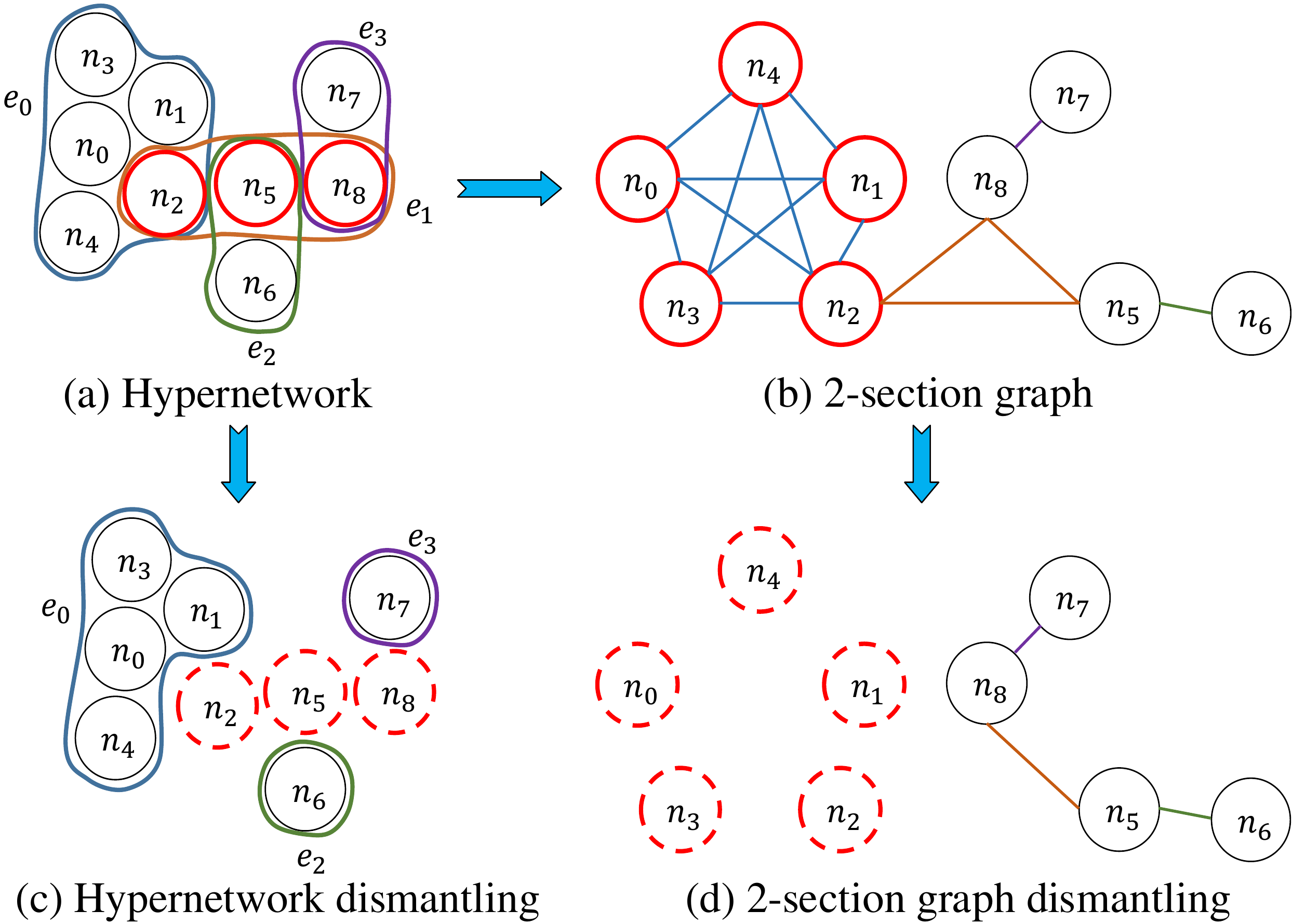}
	\caption{Hypernetwork and its 2-section graph. Nodes with red circles are critical nodes identified by dismantling methods of hypernetwork and simple network, respectively. The dash circles denote nodes have been removed.}
	\label{fig:motivation-example}
\end{figure}

\IEEEPARstart{N}{etwork} science has been widely applied to model the complicated interactions of many real-world systems such as biological \cite{biological_application1, biological_application2}, financial \cite{financial_application1, financial_application2} and social systems \cite{social_application1, social_application2}. Among many problems addressed in network science, network dismantling \cite{nd1}, i.e., finding an optimal nodes set while the removal of which will significantly degrade the connectivity of a network, is of the great importance in understanding epidemic contagion \cite{epidemic} and optimal information spreading \cite{rumour}. {Different from another graph combinatorial optimization problem of influence maximization, network dismantling focuses on the connectivity of network and utilizes only the network topology information. Its solution needs no assumption of the underlying diffusion model like linear threshold or independent cascade as influence maximization does.} Generally, conventional researches \cite{nd1, FINDER} only model pairwise interactions of such complex systems and design greedy dismantling methods according to various centrality measures on specific local or global network structures such as degree and collective influence \cite{CI}, while ignoring the ubiquitous existence of group-wise interactions and their critical roles in the formation of connectivity of these networked systems. For example, an outbreak of epidemic usually results from the group attendance of a party rather than person-to-person interactions. Moreover, heuristic methods \cite{nd,nd1} usually lack transferability, which limits their adoption in diverse real-world applications.

Fortunately, hypernetwork \cite{hypergraph} and deep learning are two promising techniques to tackle these problems. On the one hand, hypernetwork provides a generalized structure for modeling both pairwise and group-wise interactions, in which interactions among a flexible number of nodes are defined as hyperedges and a hyperedge containing two nodes is exactly an edge in simple network model. As shown in Figure \ref{fig:motivation-example} (a), the hypernetwork models group-wise interactions among nodes $n_{2}, n_{5}, n_{8}$ as hyperedge $e_1$ as well as pairwise interactions between nodes $n_{7}, n_{8}$ as hyperedge $e_3$. Although network dismantling methods for simple networks can be applied to hypernetworks with its 2-section graph \cite{2-section} as shown in Figure \ref{fig:motivation-example}(b), it will introduce too many noisy
edges, which degrades the effectiveness of existing network dismantling methods. Taking Figure \ref{fig:motivation-example} as an example of epidemic control of COVID-19, an infected person $n_2$ has attended two parties $e_0$ and $e_1$ and now what is the optimal epidemic control policy? In the hypernetwork model, nodes $\{n_{2}, n_{5}, n_{8}\}$ are critical for connectivity while in 2-section graph critical nodes are $\{n_{0}, n_{1}, n_{2}, n_{3}, n_{4}\}$. Although both dismantling strategies destroy the hypernetwork to a residual giant component of four nodes as shown in Figure \ref{fig:motivation-example} (c) and Figure \ref{fig:motivation-example} (d), hypernetwork dismantling needs lower cost, i.e., with a smaller set of removal nodes. Moreover, hypernetwork dismantling is more effective because it prevents the close contacts $n_5$ and $n_8$ from attending parties $e_2$ and $e_3$ in the future to avoid infecting nodes $n_6$ and $n_7$.

On the other hand, deep learning techniques such as representation learning and deep reinforcement learning (DRL) have been widely adopted in graph data. For example, network embedding methods such as Deepwalk \cite{deepwalk} and Hyper2vec \cite{hyper2vec} transform non-Euclidean graph data to Euclidean data by learning a low dimensional vector representation for each node in simple network and hypernetwork, respectively. Moreover, DRL has recently achieved excellent performance in tackling the graph combinatorial optimization problems by transforming them into the problem of sequential decisions. Instead of designing handcrafted heuristic methods, researches have attempted to design agent to solve various graph combinatorial optimization problems such as simple network dismantling \cite{FINDER} and influence maximization \cite{DISCO}. 

{However, existing network dismantling methods face two main challenges. The first is that existing network dismantling methods are mainly designed for simple network modeling only pairwise relations while group-wise relations are ubiquitous. Downgrading hypernetworks to simple networks will introduce noisy edges and make dismantling methods designed specifically for simple network insufficient. The second is that hypernetwork-specific dismantling is seldom studied and only several existing hypernetwork centrality measures can be utilized for hypernetwork dismantling. Moreover, these centrality measures are heuristic and handcrafted for extracting a specific topology characteristic, which not only costs much more time to design but also has limited generalization ability to various hypernetwork with different topology characteristics.}

To address these challenges, we propose \textbf{HITTER}, a \textbf{H}ypernetwork dismantling framework based on \textbf{I}nduc\textbf{T}ive hyperne\textbf{T}work \textbf{E}mbedding and deep \textbf{R}einforcement learning. Our main contributions are summarized as follows:

\begin{itemize}
	\item We propose a deep reinforcement learning based hypernetwork dismantling framework. An agent is built to practice trial-and-error dismantling tasks on large amounts of small scale synthetic hypernetworks to gain an optimal strategy and then is applied to diverse real-world hypernetwork dismantling tasks.
	\item We design a novel inductive hypernetwork embedding method to ensure transferability. Both local and global structure information is preserved with a two-level information aggregation process.
	\item We conduct extensive experiments on five real-world hypernetworks from diverse domains. The results demonstrate the effectiveness of our proposed framework.
\end{itemize}

The rest of this article is organized as follows. In Section \ref{sec:related-work}, we briefly review the important work related to this paper. In Section \ref{sec:preliminaries}, some preliminaries for the article are provided. The details of our proposed framework HITTER is described in Section \ref{sec:framework} and the experiment results are shown in Section \ref{sec:experiments}. Finally, we draw conclusions in Section \ref{sec:conclusion}.

\section{Related Work} \label{sec:related-work}
\subsection{Network Dismantling}
Network dismantling aims at finding an optimal set of nodes, the deletion of which significantly degrades the connectivity of the network. It is closely related to the robustness of networks \cite{pioro2021network, wang2021method, tian2021reinforcement}. Conventional greedy methods are usually based on local or global centrality measures. Generally the local centrality measures such as degree are easy to calculate, but their performance on network dismantling is poor. On the contrary, the global centrality measures such as betweenness are excellent on network dismantling while the high computation cost limits their application to large scale networks. Thus, some heuristic methods are proposed to achieve better performance with less time consumption. Braunstein et al. \cite{nd} proposed a three stage algorithm Min-Sum to dismantle a network through network decycling, tree breaking and cycles reinsertion. Ren et al. \cite{nd1} further took consideration of the removal cost and generalized the network dismantling problem. Since this kind of methods is weak in transferability, Fan et al. \cite{FINDER} proposed FINDER based on DRL to train an agent, which can be applied to various real-world networks. Zhao et al. \cite{zhao2020dismantling} and Li et al. \cite{li2018disintegration} extended the network dismantling problem to interdependent networks and heterogeneous combat networks, respectively. However, these methods focus on network with pairwise interaction while ignoring the group-wise interaction in real world. Therefore, we solve the dismantling problem on hypernetwork with group-wise interaction.

\subsection{Hypernetwork Embedding}
Hypernetwork embedding maps the nodes in a hypernetwork into low-dimension vectors to various downstream tasks. Huang et al. \cite{hyper2vec} proposed Hyper2vec to embed nodes in hypernetwork through random walk and skip-gram model. However, this method embeds nodes without clear tasks and cannot be trained in the way of end to end. Thus, Hyper2vec has a poor performance on various tasks. With the surge of graph neural network (GNN), researches attempt to introduce GNN into hypernetwork. Feng et al. \cite{HGNN} extended graph convolution network (GCN) to hypernetwork and proposed the HGNN model. However, the clique expansion used in HGNN introduces too many edges, which performs not well in terms of efficiency. Thus, Tadati et al. \cite{HyperGCN} proposed HyperGCN to increase the efficiency of hypernetwork embedding. Besides, the above methods focus on the static hypernetwork while the dynamic hypernetwork is more suitable to model real world. So, Jiang et al. \cite{DHGNN} proposed DHGNN to embed dynamic hypernetworks. In addition, the methods above applied GNN in hypernetwork through decomposing the hyperedges into several single edges, which lead to information loss or too much noise. The model LHCN \cite{lhcn} proposed by Bandyopadhyay et al. applied the GCN to line graph which is transformed by original hypernetwork. Generally, existing methods are transductive with limited transferability, which are insufficient to adapt the frequent change of network structure in the training stage of our framework. Therefore, we design an inductive hypernetwork embedding method to ensure transferability.

\subsection{Deep Reinforcement Learning}
Deep reinforcement learning \cite{arulkumaran2017deep,nguyen2020deep} is a promising approach to solve the high-dimension issue in reinforcement learning through the powerful representation ability of deep learning. Based on this idea, DeepMind proposed the deep Q-network (DQN) \cite{nature_DQN}. In order to alleviate the problem of data correlation, the mechanism of experience replay and a target Q-network are introduced in DQN. Based on the naive DQN, various models have been proposed. For example, Hasselt et al. presented Double DQN \cite{double_DQN} to solve the problem of $q$ value overestimating. Hausknecht et al. proposed the model dueling-DQN \cite{dueling_DQN} to improve the effectiveness of agent training. Due to its powerful ability, DRL has been widely applied to various fields such as cloud computing \cite{taskScheduling, tbsn}, IoT \cite{sdnIoT} and so on. Recently, DRL has been applied to graph data in order to address some challenging problems which are NP-hard. For example, Dai et al. proposed model S2V-DQN \cite{S2V-DQN} to solve several graph combinatorial optimization problems. In addition, Li et al. \cite{DISCO} presented a novel framework DISCO to solve the influence maximization problem. Fan et al. \cite{FINDER} proposed the framework FINDER to dismantle a simple network. As a result of the structural differences between hypernetwork and simple network, the methods above cannot be directly applied. So in this paper, we adopt DRL to solve the hypernetwork dismantling problem which is also NP-hard.

\begin{table}[]
\caption{Notations and their explanations}
\label{tab:notations}
\centering
\renewcommand\arraystretch{1.25}
\begin{tabular}{c||c}
\hline
Notation     & Explanation                                                     \\ \hline \hline
$G$            & Hypernetwork                                                     \\ \hline
$V$            & Node set                                                        \\ \hline
$E$            & Hyperedge set                                                   \\ \hline
$\bm{H}$            & Incidence matrix                                                \\ \hline
$\bm{I}$            & Identity matrix                                            \\ \hline
$GCC$          & Giant connected component                                       \\ \hline
$deg(v_{i})$, $hdeg(v_{i})$          & Degree and hyper-degree of node $v_{i}$           \\ \hline
$\bm{X}^l$, $\bm{Y}^l$         & Embeddings of node and hyperedge in the $l$-th layer                                \\ \hline
$\alpha_{v_{k},e_{j}}$        & Attention weight from node $v_{k}$ to hyperedge $e_{j}$                     \\ \hline
$\bm{W}$            & Trainable weight matrix                                        \\ \hline
$nei(e_{j})$       & Neighbors of hyperedge $e_{j}$                                       \\ \hline
$E(v_k)$         & Hyperedges which contain node $v_k$                                 \\ \hline
$r$            & Reward                                                          \\ \hline
$a$            & Node to be removed                                            \\ \hline
$s$            & The state of hypernetwork                                 \\ \hline
$q(s, a)$      & $q$ value of removing node $a$ in given state $s$                       \\ \hline
$n$           & Step length                                                     \\ \hline
$F$            & Synthetic hypernetwork generator                                \\ \hline
$\bm{\Theta}, \hat{\bm{\Theta}}$ & Parameters in Q-network and target Q-network                    \\ \hline
$\epsilon$      & Probability of $\epsilon$-greedy strategy                                  \\ \hline
$\gamma$        & Reward discount                                                 \\ \hline
$P$            & Experience pool                                                 \\ \hline
$M$            & Maximum size of experience pool                                                 \\ \hline
$C$            & Parameters copy frequency \\ \hline
$\kappa$        & Dismantling sequence                                            \\ \hline
\end{tabular}
\end{table}
\section{Preliminaries} \label{sec:preliminaries}

\begin{figure}[htbp]
	\noindent\centering
	\includegraphics[scale=0.55]{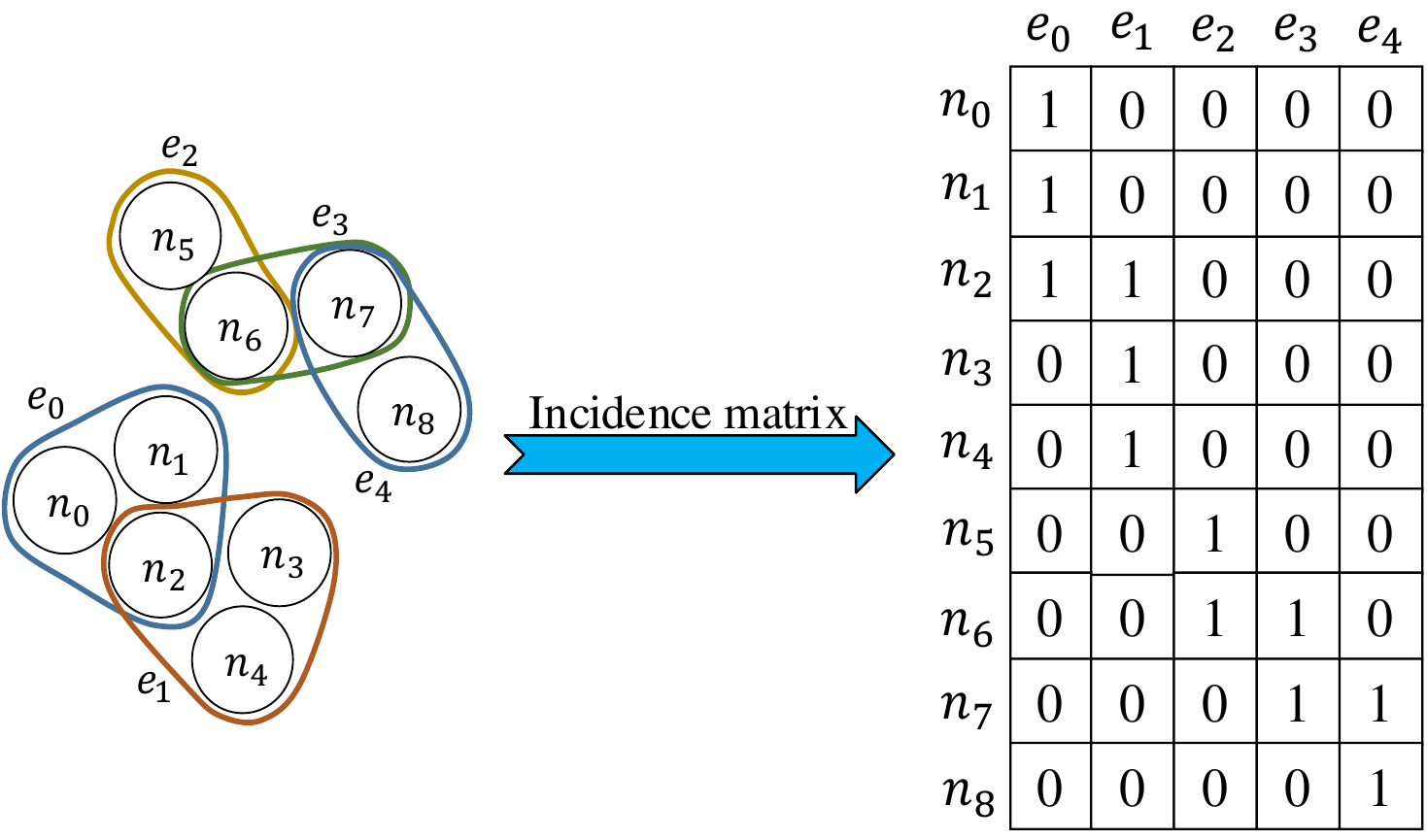}
	\caption{Hypernetwork and its incidence matrix}
	\label{fig:definitions}
\end{figure}
Hypernetwork dismantling studies the problem of finding an optimal set of nodes in a hypernetwork while the removal of which will significantly degrade the connectivity of the hypernetwork. In this section, we introduce the definitions of related concepts and problem formulation. Moreover, the notations used in this paper are summarized in Table \ref{tab:notations}. Generally, sets, vectors and matrices are denoted as upper-case letters, bold lower-case letters and bold upper-case letters, respectively.

\begin{definition}[Hypernetwork \cite{2-section}]
A hypernetwork is defined as $G = (V, E)$, where $V$ and $E$ denote the node set and the hyperedge set, respectively. Each hyperedge $e \in E$ is a subset of nodes $\{v_1,\cdots, v_k\} \subseteq V$ and the total number of nodes in a hyperedge is defined as the hyperedge size.
\end{definition}

The definition indicates that hypernetworks can model both pairwise and group-wise interactions, and simple network is a special form of it with the size of each hyperedge equal to two. Similar to simple networks, hypernetwork is usually formulated with incidence matrix defined as follows.
\begin{definition}[Incidence matrix \cite{2-section}]
The incidence matrix $\bm{H} \in \{0, 1\}^{|V| \times |E|}$ of a hypernetwork $G = (V, E)$ indicates the membership of the nodes V in the hyperedges E. Each element $H (v, e) \in H$ reflects whether the node $v$ is in the hyperedge $e$.
\end{definition}

\begin{figure*}[htbp]
	\centering
	\includegraphics[scale=0.24]{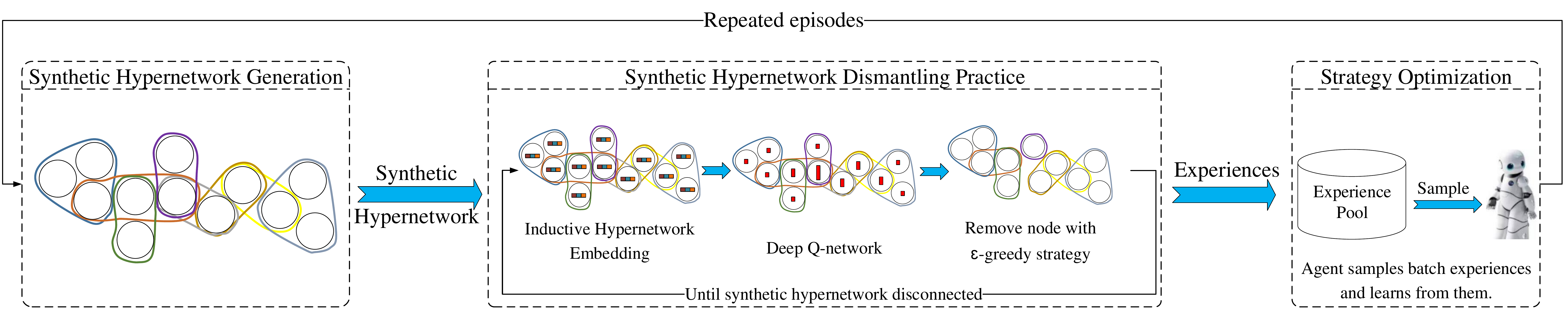}
	\caption{The framework of HITTER}
	\label{fig:framework}
\end{figure*}

Figure \ref{fig:definitions} shows an example hypernetwork and its incidence matrix. Different to nodes in simple network, nodes in hypernetwork have the properties of degree and hyper-degree which are defined as follows.
\begin{definition}[Degree \cite{hypergraph} and Hyper-degree \cite{2-section}]
The degree $d_{v_i}$ reflects how many nodes are adjacent to node $v_i$, and the hyper-degree $hd_{v_i}$ reflects how many hyperedges contain node $v_i$. The calculations of them are:
\begin{equation}
    hdeg({v_i}) = \sum_{e_j=1}^{|E|}\bm{H}(v_i, e_j)
\end{equation}
\begin{equation}
    deg({v_i}) = \sum_{v_j=1}^{|V|}\bm{HH}^{T}(v_i, v_j) - hdeg(v_{i})
\end{equation}

\end{definition}

Intuitively, the connectivity of hypernetwork is necessary in the dismantling problem. As Berge \cite{connectivity} pointed out whether a hypernetwork is connected depends on relations between hyperedges. Therefore, we define the connectivity of hypernetwork as follows.

\begin{definition}[Hypernetwork connectivity]
The connectivity of hypernetwork $G$ is defined as the ratio of nodes number in giant connected component (GCC) to total nodes number in the whole hypernetwork.
\begin{equation}
connectivity(G)=\frac{|V_{GCC}|}{|V_{G}|}
\end{equation}
where $GCC$ is a connected component of $G$ with the most hyperedges. $|V_{GCC}|$ and $|V_{G}|$ denote nodes numbers of $GCC$ and $G$, respectively.
\end{definition}

\section{Proposed Framework - HITTER} \label{sec:framework}
\subsection{Overview}

Figure \ref{fig:framework} illustrates our proposed hypernetwork dismantling framework HITTER. The framework builds and trains an agent to conduct trial-and-error hypernetwork dismantling tasks on a large amount of synthetic hypernetworks to gain an optimal strategy which can be applied to diverse real-world hypernetworks. It consists of three main components: 1) \textbf{synthetic hypernetwork generation}, which generates small synthetic hypernetworks according to a hypernetwork generation model; 2) \textbf{synthetic hypernetwork dismantling practice}, which adopts inductive hypernetwork embedding and the Q-network to dismantle the synthetic hypernetwork and then saves the experiences from the dismantling process into experience pool; 3) \textbf{strategy optimization}, which optimizes the dismantling strategy according to experiences sampled from experience pool. We will explain each component in detail in the following subsections.

\subsection{Synthetic Hypernetwork Generation}

HITTER formulates hypernetwork dismantling as a node sequence decision problem on hypernetworks and solves it with DRL models which usually need plenty of training data. In order to feed enough training data to train a DRL model, HITTER first employs generative hypernetwork models such as HyperPA \cite{hyperpa} and HyperFF \cite{hyperff} to generate small synthetic hypernetworks. HyperPA generates hypernetworks according to a predefined degree distribution while HyperFF can generate hypernetworks with various degree distributions by introducing two flexible parameters, i.e., the burning and expanding probability. In the HITTER framework, an agent needs to explore on diverse hypernetworks to learn a more flexible dismantling strategy. So HyperFF is chosen as the synthetic hypernetwork generator.

\subsection{Synthetic Hypernetwork Dismantling Practice}

\begin{figure*}[htbp]
	\centering
	\includegraphics[scale=0.38]{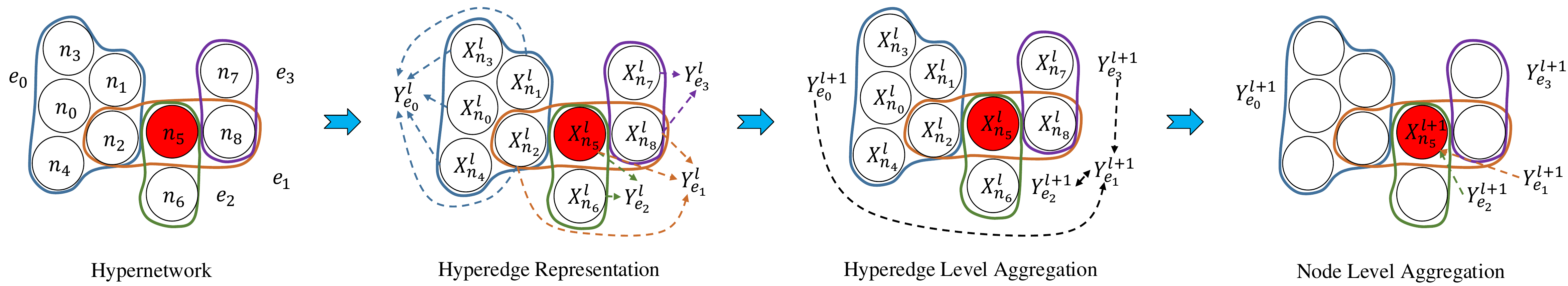}
	\caption{The process of information aggregation in HyperSAGE}
	\label{fig:hypersage}
\end{figure*}

In order to apply the DRL method on non-Euclidean hypernetwork data, hypernetwork embedding is needed to transform a hypernetwork to a low dimensional vector space. Moreover, the agent in HITTER is trained on synthetic hypernetworks to ensure the transferability to real-world hypernetworks. Therefore, the hypernetwork embedding should be inductive. Inspired by GraphSAGE \cite{graphsage}, we design an inductive two-level hypernetwork embedding method, \textbf{HyperSAGE} to aggregate information from both the hyperedge level and the node level iteratively. The whole process of HyperSAGE is depicted in Figure \ref{fig:hypersage}.

\textbf{Hyperedge level aggregation} aggregates information for a hyperedge from its neighbor hyperedges. Intuitively, hyperedge should be first represented as a summarization of all nodes in it. However, as different nodes contribute differently, the attention mechanism is adopted as follows:

\begin{equation}
\label{eq:nodes-merge}
\bm{Y}^{l}_{e_{i}} = \sum_{v_{k} \in e_{i}}\alpha_{v_{k},e_{i}}\bm{X}_{v_{k}}^{l}
\end{equation}
\begin{equation}
\alpha_{v_{k},e_{i}} = \frac{\exp(\bm{W}_{1}\bm{X}_{v_{k}}^{l})}{\sum_{v_{p} \in e_{i}}\exp(\bm{W}_{1}\bm{X}_{v_{p}}^{l})}
\end{equation}
where $\bm{X}^{l}$ and $\bm{Y}^{l}$ denote the node and the hyperedge representations in the $l$-th layer, respectively. $\bm{W}_{1} \in \mathbb{R}^{1 \times d_{l}}$($d_{l}$ is the dimension of nodes and hyperedges representation in the $l$-th layer) is the model parameter. And $v_{k}$, $e_{i}$ denote the node $k$ and the hyperedge $i$, respectively.

Then representations from both self and neighbor hyperedges are aggregated together as the new representation for each hyperedge with Equations (\ref{eq:hyperedge-representation-neighbor}) and (\ref{eq:hyperedge-representation-aggregation}).
\begin{equation}
\label{eq:hyperedge-representation-neighbor}
\bm{Y}^{l}_{nei(e_{i})} = \sum_{e_{j} \in nei(e_{i})}\frac{1}{\sqrt{|nei(e_{i})|}\sqrt{|nei(e_{j})|}}\bm{Y}^{l}_{e_{j}}
\end{equation}

\begin{equation}
\label{eq:hyperedge-representation-aggregation}
\bm{Y}_{e_{i}}^{l + 1} = f(\bm{W}_{4}^{T}[\bm{W}_{2}\bm{Y}^{l}_{nei(e_{i})}||\bm{W}_{3}\bm{Y}^{l}_{e_{i}}])
\end{equation}
where $nei(e_{i})$ denotes hyperedge neighbors of $e_{i}$ which share common nodes, $\bm{W}_{2},\bm{W}_{3} \in \mathbb{R}^{d_{l + 1} \times d_{l}}$ and $\bm{W}_{4} \in \mathbb{R}^{2d_{l + 1} \times d_{l + 1}}$ are the parameters. $||$ means concatenation operation, and the activation function $f$ is specified as ReLU in this paper.

\begin{algorithm}[htbp]
  \caption{HyperSAGE}
  \label{algo:hypersage}
  \begin{algorithmic}[1]
    \REQUIRE Hypernetwork incidence matrix $\bm{H}$, Embedding dimension $d$, Number of layers $L$, and Initialized feature of nodes $\bm{X}^{0}$
    \ENSURE Node embeddings $\bm{X}$ and hyperedge embeddings $\bm{Y}$
    \FOR{$l$ = 0 to $L - 1$}
      \STATE Merge node embeddings into hyperedge embeddings $\bm{Y}$ according to Equation (\ref{eq:nodes-merge})
      \STATE Perform hyperedge level aggregation and get hyperedge embeddings $\bm{Y}^{l + 1}$ with Equations (\ref{eq:hyperedge-representation-neighbor}) and (\ref{eq:hyperedge-representation-aggregation})
      \STATE Perform node level aggregation and get node embeddings $\bm{X}^{l + 1}$ according to Equation (\ref{eq:node-representation})
    \ENDFOR
    \STATE  Set $\bm{X}=\bm{X}^{L}$, $\bm{Y}=\bm{Y}^{L}$
    \RETURN $\bm{X}$, $\bm{Y}$
  \end{algorithmic}
\end{algorithm}
\textbf{Node level aggregation} aggregates the representation of a node from the hyperedges containing it, which is formulated in Equation (\ref{eq:node-representation}) similar to the hyperedge level aggregation.
\begin{equation}
\label{eq:node-representation}
\bm{X}_{v_{k}}^{l + 1} = f(\bm{W}_{7}^{T}[\sum_{e_{i} \in E(v_{k})}\bm{W}_{5}\bm{Y}_{e_{i}}^{l + 1}||\bm{W}_{6}\bm{X}_{v_{k}}^{l}])
\end{equation}
where $E(v_{k})$ is a set of hyperedges containing node $v_{k}$. $\bm{W}_{5} \in \mathbb{R}^{d_{l + 1} \times d_{l + 1}},\bm{W}_{6} \in \mathbb{R}^{d_{l + 1} \times d_{l}}$ and $\bm{W}_{7} \in \mathbb{R}^{2d_{l + 1} \times d_{l + 1}}$ are the parameters. 

Node level aggregation preserves local structure information for a node from hyperedges containing it. The two-level aggregation process runs iteratively in each layer, and multiple layers can be chained to preserve global information. The detailed procedure of HyperSAGE is described in Algorithm \ref{algo:hypersage}.

Once the embeddings are obtained, the agent maps the process of hypernetwork dismantling into the decision process in DRL: 1) the state is the residual hypernetwork, the embedding of which is obtained by inserting a virtual node which only receives information from all hyperedges while not influencing the aggregation process during the inductive hypernetwork embedding; 2) the action is a node to be removed; and 3) the reward is related to the connectivity of residual hypernetwork and is formulated as Equation (\ref{eq:reward}), where $G'$ is the residual hypernetwork. Actually, the reward is a form of punishment to prevent the agent from abusing limited budget. In other words, to significantly destroy the connectivity of a hypernetwork within $K$ removal nodes, each action must be optimally decided to reduce punishment.
\begin{equation}
\label{eq:reward}
r = - connectivity(G')
\end{equation}

The $\epsilon$-greedy strategy is adopted to balance exploration and exploitation during the dismantling process of a synthetic hypernetwork. Specially, the agent selects and removes a node with the highest $q$ value by a Q-network with a probability $1-\epsilon$ or a random node otherwise. The Q-network is implemented using a multi-layer perceptron defined in Equation (\ref{eq:qvalue}):
\begin{equation}
\label{eq:qvalue}
q(s, a) = \bm{W}_{8}^{T}f(\bm{X}_{s}^{T}\bm{X}_{a}\bm{W}_{9})
\end{equation}
where $\bm{W}_8$ and $\bm{W}_{9} \in \mathbb{R}^{d_{L} \times 1}$ are the parameters ($d_{L}$ is the node and hyperedge embeddings dimension in the final layer). $\bm{X}_{a} \in \mathbb{R}^{1 \times d_{L}}$ and ${\bm{X}_{s}} \in \mathbb{R}^{1 \times d_{L}}$ denote the representation of the action and current hypernetwork state in the final layer, respectively. $q(s, a)$ is the predicted reward of removing node $a$ in given state $s$, which reflects the importance of $a$.

An entire dismantling practice on one synthetic hypernetwork, i.e., an episode, terminates until the synthetic hypernetwork is completely disconnected and a decision sequence $S$ is obtained, from which experience can be extracted as a four-element tuple by taking delayed rewards into account with $n$-step Q-learning.
\begin{equation}
\label{eq:state-action-sequence}
S = ({s_{0}, a_{0} ,r_{0}, \cdots, s_{T - 1}, a_{T - 1}, r_{T - 1}, S_{T}})
\end{equation}
\begin{equation}
\label{eq:experience}
experience = (s_{t}, a_{t}, r_{t, t + n}, s_{t + n})
\end{equation}
where $n$ is the step length, and $r_{t, t + n} = \sum_{j=t}^{t+n}r_{j}$ denotes accumulated rewards. Intuitively, each episode will contribute more than one experience.

\subsection{Strategy Optimization}

In order to optimize the dismantling strategy, batch experiences are sampled to update the agent with the optimization objective considering both deep Q-network and hypernetwork reconstruction. For each experience, the loss of deep Q-network aims at minimizing reward error between prediction and ground-truth. Thus, the loss of this part is defined as the mean-square-loss between the predicted $q$ value and actual reward according to Bellman Equation, which is shown in Equation (\ref{eq:loss-dqn}),

\begin{equation}
\label{eq:loss-dqn}
L_{Q} = {(r_{t,t+n}+\gamma\max\limits_{a}\hat{q}(s_{t+n},a)-q(s_t,a_t))}^2
\end{equation}
where $\gamma$ is the reward discount, which balances the importance of future and current rewards, and $\hat{q}(s_{t+n},a)$ is the $q$ value given by target Q-network.

In addition, the hypernetwork reconstruction loss is designed to preserve structure information of the hypernetwork by restraining hyperedge embeddings, and the loss of this part is shown in Equation (\ref{eq:loss-embedding}).
\begin{equation}
\label{eq:loss-embedding}
\begin{aligned}
L_{E} &= \sum_{e_i \in E}\sum_{e_j \in nei(e_i)}||\bm{Y}_{e_i} - \bm{Y}_{e_j}||^2_2 \\
&= 2 \times tr(\bm{Y}^{T}(\bm{I}-\bm{H}^{T}\bm{H})\bm{Y})
\end{aligned}
\end{equation}
where $tr$ denotes the trace of matrix, $\bm{Y}$ is the hyperedge embeddings in given hypernetwork state of experience. $\bm{I}$ and $\bm{H}$ are identity matrix and hypernetwork incidence matrix, respectively.

The total loss is a combination of $L_{Q}$ and $L_{E}$, and parameter $\alpha$ is introduced to balance them.
\begin{equation}
L = L_{Q} + \alpha L_{E}
\label{eq:total_loss}
\end{equation}

With repeatedly gathering experiences and learning from them, the agent updates its hypernetwork dismantling strategy continuously. Finally, the agent can learn an optimal strategy which is suitable for real-world hypernetwork dismantling. The detailed procedure is described in Algorithm \ref{algo:hitter-training-stage}.

Specially, lines 1-3 denote the initialization of the agent's parameters and the experience pool. Then, a synthetic hypernetwork is generated and supplied to the agent in line 5. In lines 6-14, the agent does trial-and-error dismantling on this synthetic hypernetwork and experiences are gathered and saved into the experience pool. Finally, in lines 15-17, the agent samples a batch of experiences from the experience pool and updates the dismantling strategy from them. After several episodes, the agent learns a near optimal hypernetwork dismantling strategy, which can be applied to the dismantling of real world hypernetworks.

\subsection{Time Complexity}
The time complexity of HyperSAGE is relevant to the layer number $L$. Intuitively, there are three parts in each layer. The first part is the representation of hyperedges by the attention mechanism, and the time complexity of which is $O(|E|d)$ ($|E|$ denotes the number of 
\begin{algorithm}
  \caption{HITTER}
  \label{algo:hitter-training-stage}
  \begin{algorithmic}[1]
    \REQUIRE Synthetic hypernetwork generator $F$, max episode number $N$, multi step length $n$, {max size of experience pool $M$}, exploration probability $\epsilon$, and target Q-network copy frequency $C$
    \ENSURE Agent parameters $\bm{\Theta}$
    \STATE Initialize experience pool $P$ with max size $M$
    \STATE Initialize agent parameters $\bm{\Theta}$ randomly
    \STATE Initialize target Q-network parameters $\hat{\bm{\Theta}} = \bm{\Theta}$
    \FOR{$episode$ = 1 to $N$}
      \STATE Generate a synthetic hypernetwork $G$ by $F$
      \STATE Initialize state sequence $S$ = ()
      \WHILE{$G$ is connected}
        \STATE Embed nodes and state $s$ using HyperSAGE
        \STATE Select $a=\left\{
        \begin{aligned}
        &\rm{random\ node}, \ \ \ \ \ \rm{with\ probability} \ \epsilon \\
        &\arg\max\limits_{a}q(s, a), \ \ \ \ \ \ \ \  \ \rm{otherwise}
        \end{aligned}
        \right.$
        \STATE Remove node $a$ and get reward $r$
        \STATE Insert $(s, a, r)$ into $S$
      \ENDWHILE
	   \STATE Insert terminal state $s_{T}$ into $S$
      \STATE Extract experiences from $S$ according to Equation (\ref{eq:experience}) and save them into $P$
      \STATE Sample batch experiences from $P$ randomly
      \STATE {Calculate the total loss $L$ of batch experiences according to Equation (\ref{eq:total_loss})}
      \STATE Update $\bm{\Theta}$ using Stochastic Gradient Descent {according to total loss $L$}
      \STATE Update target Q-network parameters $\hat{\bm{\Theta}} = \bm{\Theta}$ every $C$ episodes
  \ENDFOR
  \RETURN Agent parameters $\bm{\Theta}$
  \end{algorithmic}
\end{algorithm}
hyperedges and $d$ is the embedding dimension). Then, it also takes $O(|E|d)$ to conduct the hyperedge level aggregation. In the node level aggregation, the time complexity is relevant to the number of node $|V|$ and embedding dimension $d$. Therefore, the total time complexity of HyperSAGE is $O((|V| + 2|E|)Ld)$. For Algorithm \ref{algo:hitter-training-stage}, it is obvious that HITTER consists synthetic hypernetwork generation, synthetic hypernetwork dismantling and parameters updating. Actually, in the part of synthetic hypernetwork dismantling (i.e., the inner loop in Algorithm \ref{algo:hitter-training-stage}), it is hard to be determined how many steps are needed to ensure that the hypernetwork is disconnected. However, the complexity of both residual hypernetwork embedding and node removal depends on the number of steps. Thus, it is unable to determine the time complexity of the part of hypernetwork dismantling practice, which further causes that the complexity of HITTER also cannot be determined.

\section{Experiments} \label{sec:experiments}

{In this section, we evaluate the effectiveness of HITTER with experimental analysis. In particular, we aim to answer the following questions:
\begin{itemize}
    \item \textbf{RQ1} How does HITTER perform compared to the state-of-the-art hypernetwork dismantling methods on different real-world hypernetwork?
    \item \textbf{RQ2} Do different types of hypernetwork embedding methods affect the performance and stability of HITTER?
    \item \textbf{RQ3} How do different hyperparameters affect the performance of HITTER?
    \item \textbf{RQ4} Can HITTER be applied effectively to real-world applications?
\end{itemize}
}

\subsection{Experiment Datasets and Settings} \label{sec:experiment-dataset-settings}
\subsubsection{Datasets} We evaluate the performance of our proposed HITTER framework on five real-world hypernetworks from different domains, i.e., {Cora-co-authorship}, Citeseer, Pubmed, MAG and NDC. The former three datasets are from \cite{HyperGCN}, and the latter two datasets are from \cite{data1} and \cite{data2}. Since the network dismantling focuses on the scale of GCC, we only select GCCs for all datasets and perform hypernetwork dismantling on them. 
\begin{table}[htbp]
	\caption{The statistics of datasets}
	\label{tbl:dataset-statistics}
	\centering
	\renewcommand\arraystretch{1.5}
 	\setlength{\tabcolsep}{0.8mm}{
	\begin{tabular}{lrrrrr}
		\toprule
		Datasets & {Cora-co-authorship} & Citeseer & MAG & NDC & Pubmed \\
		\midrule
		{\# Nodes} & 1,676 & 1,019 & 1,669 & 3,065 & 3,824 \\
		
		{\# Hyperedges} & 463 & 626 & 784 & 4,533 & 5,432 \\

		{Avg. hyper-degree} & 1.66 & 2.23 & 1.59 & 13.57 & 7.45 \\

		{Avg. hyperedge size} & 6.00 & 3.63 & 3.38 & 9.17 & 5.25 \\
		\toprule
	\end{tabular}}
\end{table}
{Specifically, for each of the dataset from the original literature a hypernetwork is first constructed with hyperedge built from the corresponding group-wise relation. Then the number of hyperedges for each connected component is calculated and the one with the biggest size, i.e., the Giant Connected Component is selected for the experiment.} The statistics of the datasets are summarized in Table \ref{tbl:dataset-statistics}. Detailed descriptions of these datasets are shown as follows:

\begin{table*}[]
\caption{The overall performance (Bold and Underline denote the best and the second best performed models, respectively.)}
\label{tbl:overall-performacne}
\centering
\renewcommand\arraystretch{1.5}
\begin{threeparttable}
\setlength{\tabcolsep}{2.0mm}{
    \begin{tabular}{ccccccccccc}
    \toprule
    Datasets & HD     & HDA & CI     & GND    & FINDER    & HHD    & HHDA & {SubTSSH}   & HITTER$_{trans}$ & HITTER \\ \midrule
    {Cora-co-authorship}     & 0.1591 & 0.1564 & 0.1181  & 0.1111 & 0.4068 & 0.0996 & \underline{0.0977} & {0.1267} & 0.3292        & \bf{0.0792$^{{*}}$} \\
    Citeseer & 0.1167 & 0.0930 & 0.0915  & 0.2528 & 0.1109 & 0.0815 & \underline{0.0788} & {0.0965} & 0.2416        & \bf{0.0607$^{{*}}$} \\
    MAG      & 0.0363 & 0.0261 & 0.0238  & 0.0335 & 0.0410 & \underline{0.0191} & 0.0195 & {0.0226} & 0.0757        & \bf{0.0130$^{{*}}$} \\
    NDC      & 0.2824 & 0.2608 & 0.2623  & 0.4372 & 0.4804 & 0.2561 & \underline{0.2374} & {0.2666} & 0.3566        & \bf{0.2209$^{{*}}$} \\
    Pubmed   & 0.4279 & 0.3933 & 0.3930  & \underline{0.3606} & 0.4809 & 0.4104 & 0.3831 & {0.4038} & 0.4654        & \bf{0.3529$^{{*}}$} \\ \toprule
    \end{tabular}}
    \begin{tablenotes}
        \item {$*$ means the result is significantly better than the second best method according to $t$-test at level of 0.05 for the $p$ value.}
    \end{tablenotes}
\end{threeparttable}
\end{table*}

\begin{itemize}
	\item \textbf{{Cora-co-authorship}} The {Cora-co-authorship} dataset contains publications which belong to the field of machine learning and each paper is written by several authors. We construct the hypernetwork with authors as nodes and co-author relations as hyperedges.
	\item \textbf{Citeseer} The Citeseer dataset is also a citation network like {Cora-co-authorship}. We construct the corresponding hypernetwork in a way similar to {Cora-co-authorship}.
	\item \textbf{MAG} The MAG dataset contains publications marked with the "History" tag in the Microsoft Academic Graph. Since it also reflects the co-author relations between authors, the hypernetwork is constructed similar to the {Cora-co-authorship} and Citeseer datasets.
	\item \textbf{Pubmed} The Pubmed dataset contains papers about diabetes. Nodes and edges in this dataset denote papers and the citation relation between papers, respectively. So, we construct a hypernetwork with articles as nodes and references in an article as hyperedges.
	\item \textbf{NDC} The NDC dataset is a drug-substance hypernetwork with substances as nodes and co-existing relations in a drug as hyperedges. One drug is consisted of multiple substances and one substance can contribute multiple drugs.
\end{itemize}

\subsubsection{Baselines} We compare HITTER with several baselines and the brief descriptions are shown as follows:
\begin{itemize}
	\item \textbf{Highest Degree (HD)} HD removes nodes according to their degree. In the first, degree of each node is calculated. Then, the node with the highest degree is removed in each step.
	\item \textbf{Highest Degree Adaptive (HDA)} HDA is the adaptive version of HD. As a result of structure change after each node removal step, the degree of residual nodes are recalculated in HDA.
	\item \textbf{Highest Hyper-Degree (HHD)} HHD removes nodes according to their hyper-degree. The hyper-degree of each node is calculated and nodes are removed in a descending order of hyper-degree.
	\item \textbf{Highest Hyper-Degree Adaptive (HHDA)} HHDA is the adaptive version of HHD. Similar to the HDA, HHDA recalculates hyper-degree after each removal.
	\item \textbf{Collective Influence (CI) \cite{CI}} CI of a node is calculated through its degree and the degree sum of neighbors within a constant hop, which can be used to reflect the node's reachability to other nodes. Thus, node with the highest CI value will be removed in each step.
	\item \textbf{GND \cite{nd1}} GND reduces the scale of GCC in a network through partitioning it into several sub-networks. Moreover, it adopts the policy of nodes reinsertion to optimize the dismantling set.
	\item \textbf{FINDER \cite{FINDER}} FINDER maps the network dismantling into a sequential decision problem. Through graph embedding and DQN, FINDER calculates the nodes' $q$ values which can be viewed as the contributions to dismantling and the node with the highest $q$ value is removed in each step.
	\item \textbf{{SubTSSH \cite{SubTSSH}}} {SubTSSH is a hypernetwork specific method for minimum target set selection which is similar and applicable to hypernetwork dismantling. The nodes removal and influence propagation are conducted iteratively until all nodes are removed.}
	\item \textbf{HITTER$_{trans}$} HITTER$_{trans}$ is a variant of HITTER. To show the effectiveness of the inductive hypernetwork embedding against transductive ones, HITTER$_{trans}$ replaces our designed inductive hypernetwork embedding method HyperSAGE with the transductive HGNN \cite{HGNN}.
\end{itemize}

Intuitively, the baselines above can be divided into three classes. The first class is centrality-based greedy methods (i.e., {HD, HDA, HHD, HHDA, CI and SubTSSH}). In this kind of methods, the corresponding centralities of nodes are calculated and the node with the highest centrality will be removed greedily. The GND belongs to the second class which is based on graph partition. It partitions a network into several sub-networks and achieves the purpose of network dismantling. Lastly, the FINDER solves the simple network dismantling through DRL. Besides, these methods which are based on simple network (i.e., HD, HDA, CI, GND, FINDER) are adopted to hypernetwork dismantling by transforming the original hypernetwork into its 2-section graph.

\begin{figure*}
	\centering
	\subfigtopskip=0pt
	\subfigure[{Cora-co-authorship}]{
		\includegraphics[scale=0.35]{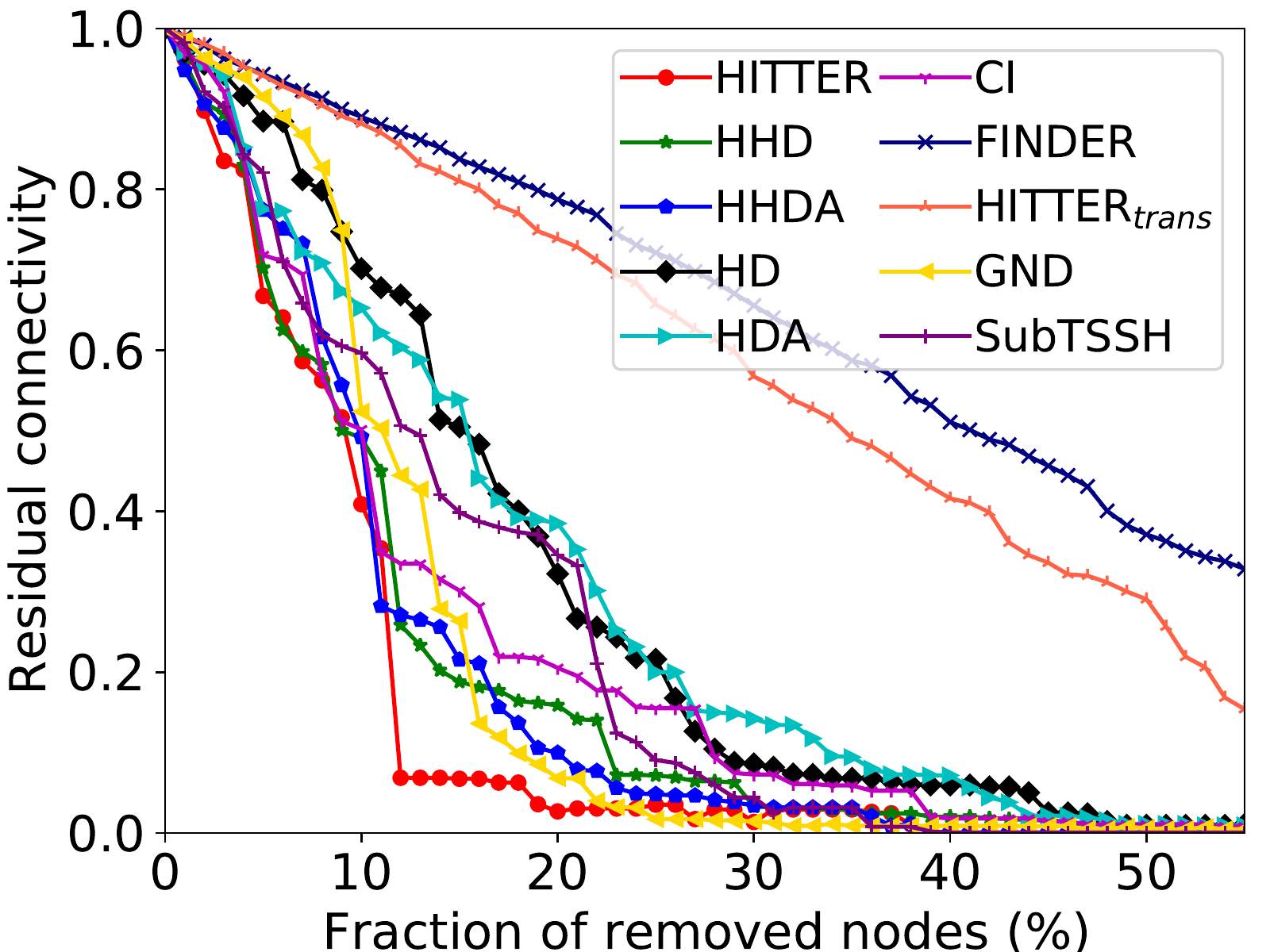}
		\label{4a}
	}
	\subfigure[Citeseer]{
		\includegraphics[scale=0.35]{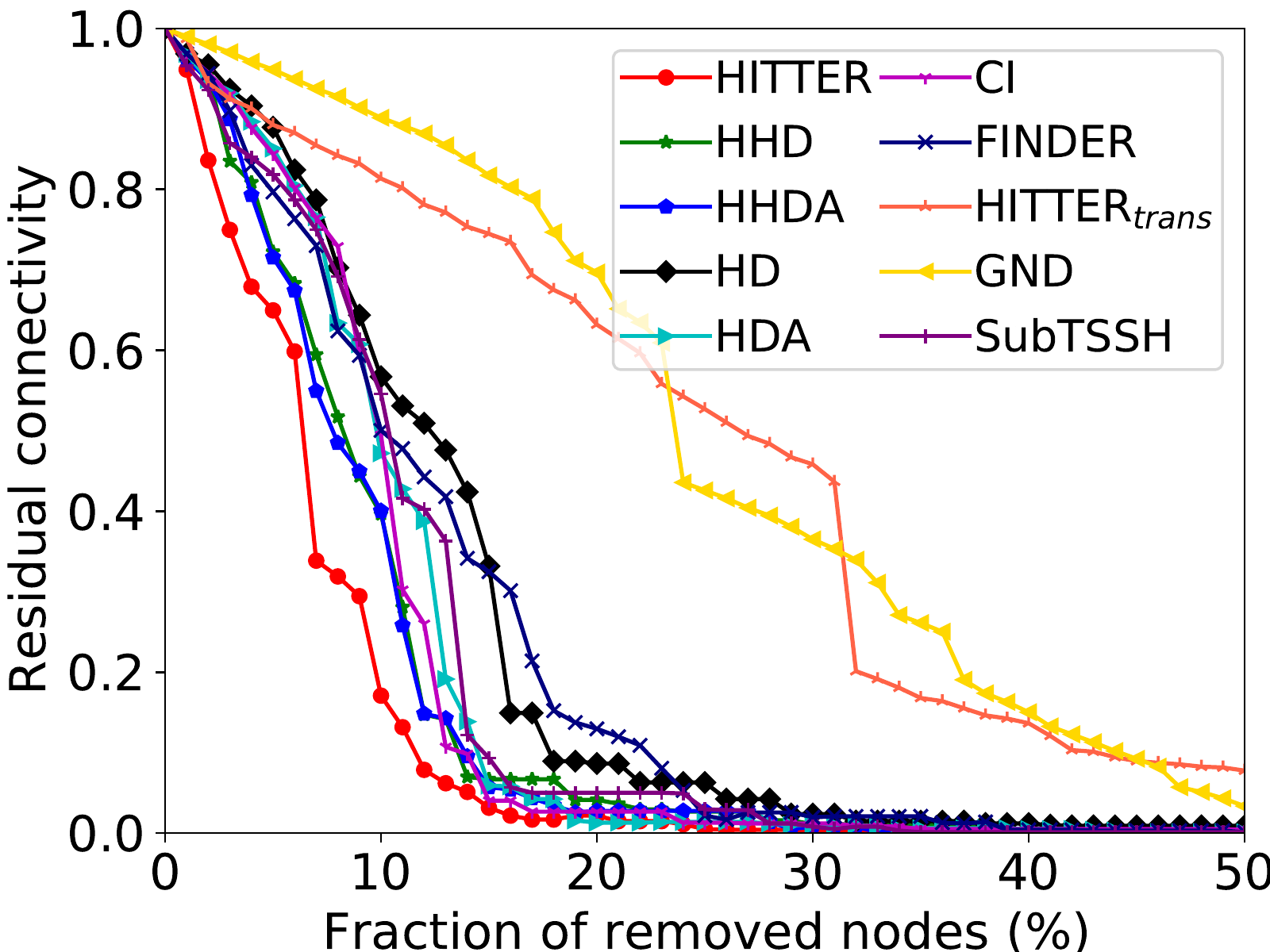}
		\label{4b}
	}
	\subfigure[MAG]{
		\includegraphics[scale=0.35]{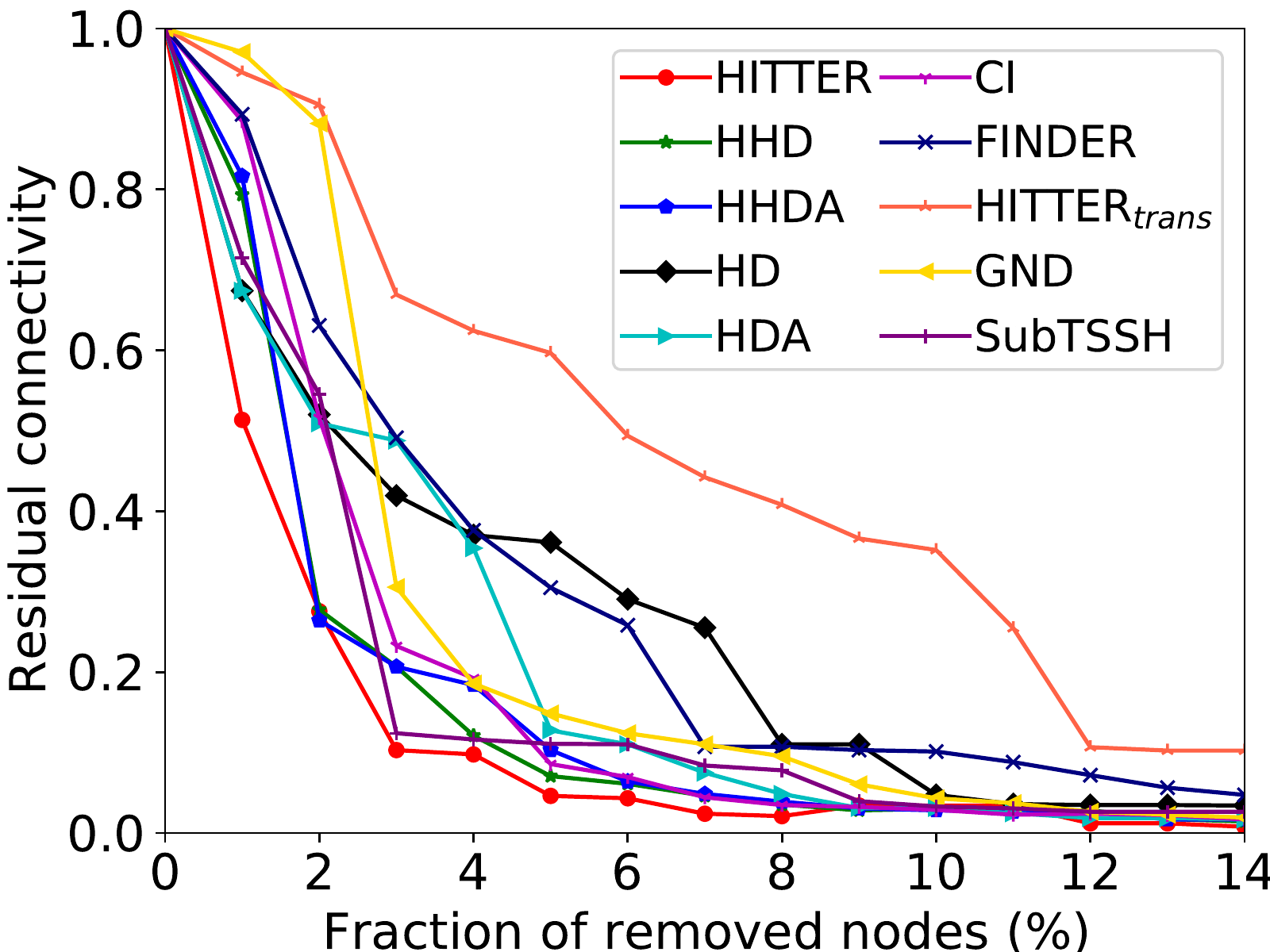}
		\label{4c}
	}
	\subfigure[NDC]{
		\includegraphics[scale=0.35]{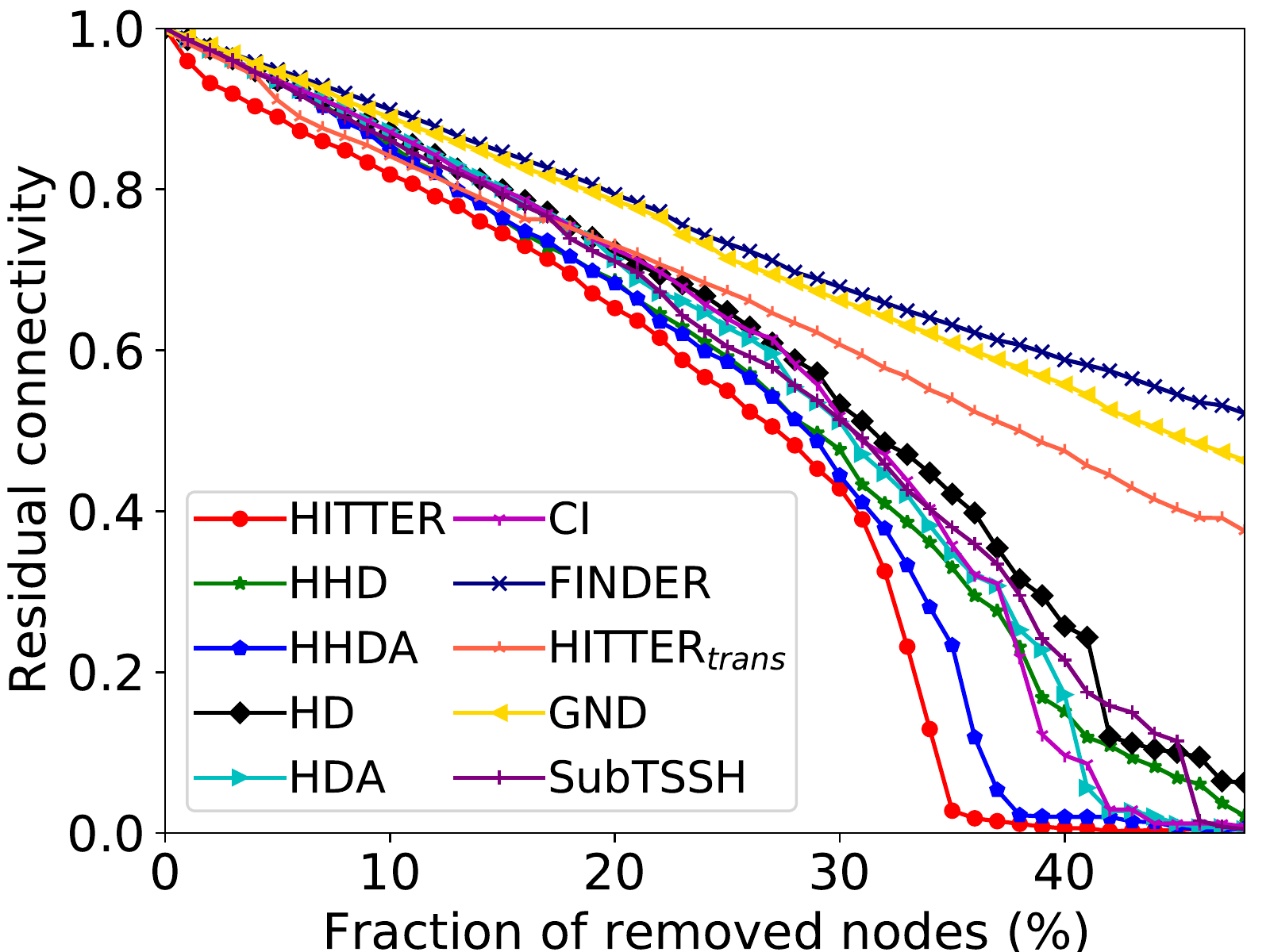}
		\label{4d}
	}
	\subfigure[Pubmed]{
		\includegraphics[scale=0.35]{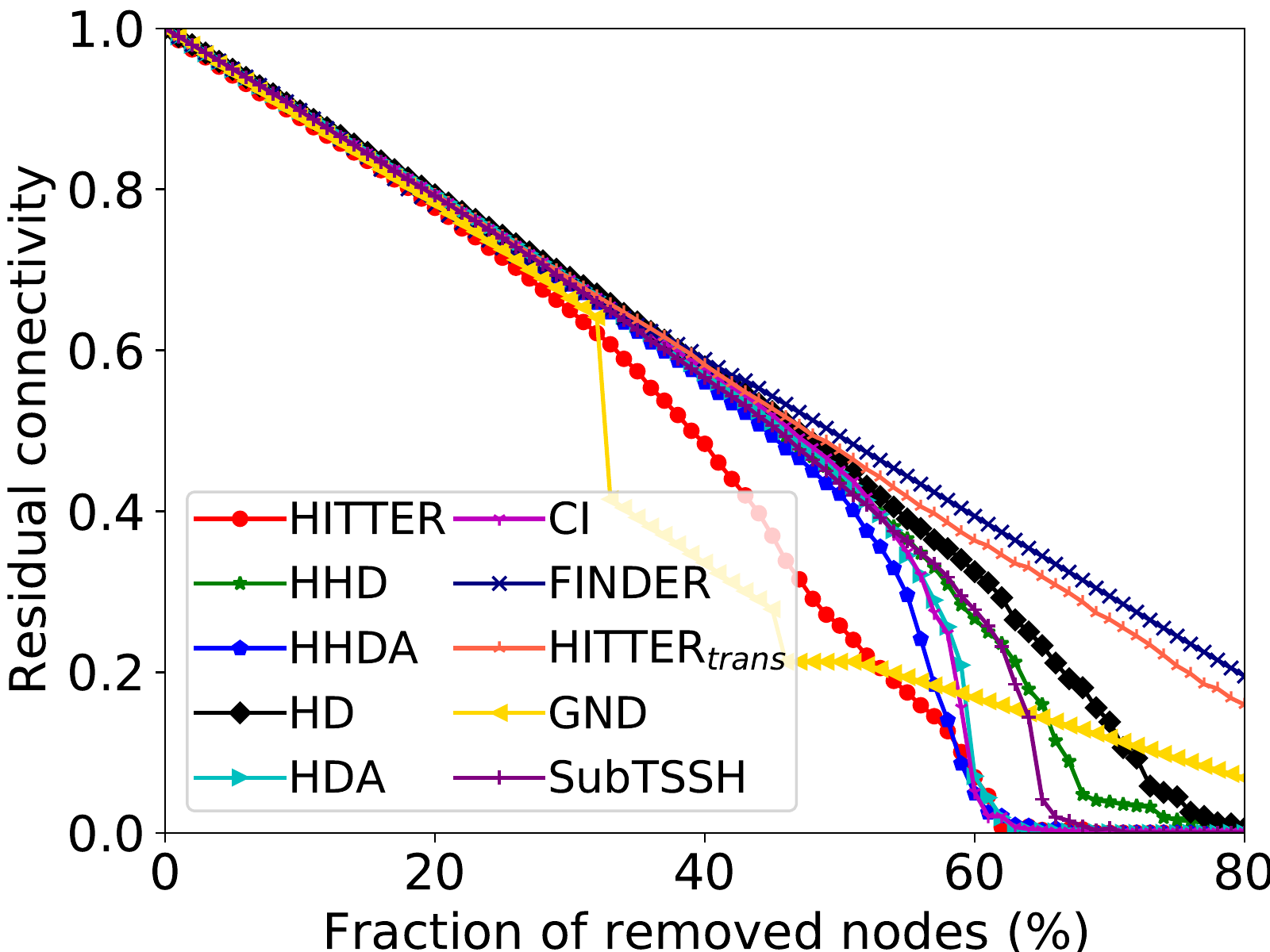}
		\label{4e}
	}
	\caption{Detailed ANC curve}
	\label{fig:detailed-anc-curve}
\end{figure*}

\subsubsection{Metrics} The accumulated normalized connectivity ($ANC$) \cite{ANC} is adopted to evaluate the dismantling performance of our method and baselines on each dataset. Intuitively given a node remove sequence $\kappa = \{v_1, v_2, \cdots, v_K\}$, the $ANC$ value on hypernetwork $G$ is calculated as follows:

\begin{equation}
ANC(\kappa)= \frac{1}{K}\sum_{k=1}^{K}\frac{connectivity(G\backslash\{v_1,v_2,\cdots,v_k\})}{connectivity(G)}
\end{equation}
where $K$ is the maximun nodes removal number and $G\backslash\{v_1,v_2,\cdots,v_k\}$ denotes a new hypernetwork after removing nodes set $\{v_1,v_2,\cdots,v_k\}$ from $G$.

\subsubsection{Implementation} \label{sec:implementation} Synthetic hypernetworks with 30-50 nodes are generated by HyperFF \cite{hyperff} with the probabilities of both burning and expanding set to 0.1. Moreover, we set node initialized features $\bm{X}^{0}$ as one-vector due to the lack of them in synthetic hypernetwork. For hypernetwork embedding, the number of layers and embedding dimension are set to 3 and 64, respectively. {Inspired by \cite{FINDER}}, the future discount $\gamma$, the length of multi-step $n$, {the experience pool size $M$} and $\epsilon$ for the $\epsilon$-greedy strategy are set to 0.99, 5, {50000} and 0.05, respectively, and 50 synthetic hypernetworks are firstly generated as validation hypernetworks. {According to \cite{FINDER}}, we conduct 1000 episodes without parameters updating in order to ensure there are enough experiences in experience pool for sampling. The max episode number $N$ is set to 100000, and the frequency of parameters copying for target Q-network is set to 1000. During training, the agent is validated on these validation hypernetworks every 50 episodes and the agent with a minimal ANC value is chosen to validate the effectiveness on the five real-world datasets, during which $1\%$ nodes are removed in each step until all nodes are removed.

\subsection{Experimental Results {(RQ1)}}
% and detailed ANC curve  and Figure \ref{fig:detailed-anc-curve}
{To answer RQ1, we compare HITTER against baselines and t}he overall performance on the five real-world hypernetworks is shown in Table \ref{tbl:overall-performacne}. The results demonstrate that our proposed framework HITTER outperforms all the baselines which indicates HITTER can learn a more optimal dismantling strategy than both greedy methods and learning based dismantling methods for simple network. Moreover, the detailed ANC curve on each dataset is also shown in Figure \ref{fig:detailed-anc-curve}. Generally, the removal of only a small portion (e.g. nearly $10\%$ in both {Cora-co-authorship} and Citeseer, $5\%$ in MAG) of nodes identified by HITTER will significantly degrade the connectivity of the hypernetworks. However, nearly one third of nodes for NDC and more than a half of nodes for Pubmed are needed. According to Table \ref{tbl:dataset-statistics}, we can see that the average hyper-degree of both NDC and Pubmed are extremely large than the others, indicating their strong resilience to attacks.

Comparing the performance of dismantling methods specially designed for hypernetwork (i.e., {HHD, HHDA and SubTSSH}) and those methods designed for simple network dismantling (i.e., HD, HDA, CI, GND and FINDER), it can be found that the former methods are often better than the latter ones. This indicates that although it simplifies the problem by transforming the hypernetwork to its 2-section graph and applying simple network dismantling methods, the performance is often poor. Actually, the transformation often introduces too many noisy edges, which drops the effectiveness of these methods and the motivation example in Figure \ref{fig:motivation-example} has also intuitively illustrated this. Specially, FINDER, which performs excellently on simple networks has a performance even worse than other simple network centrality-based methods. We believe that the BA model \cite{ba}, a simple network generation model used in FINDER, is not suitable to capture the properties of hypernetworks. Thus it is important to develop specifically designed dismantling methods for hypernetworks.

In addition, HHDA usually has better performance than HHD. HHD calculates the hyper-degree of each node only once at the initial state of the hypernetwork and decides the optimal node set for hypernetwork dismantling while HHDA attempts to remove a batch of nodes each step and then observes the state of the hypernetwork to decide the next step. The inspires us to adopt an adaptive design paradigm in future hypernetwork dismantling methods.

\subsection{Model Analysis {(RQ2)}}

In this section, we analyze the impact of different hypernetwork embedding methods on the performance of HITTER {to answer RQ2}. {As the representation of the states is always a key factor for deep reinforcement learning-based models, the component of hypernetwork embedding which extracts state information for the DQN in the framework of HITTER will surely affect the performance. In the analysis,} we replace the HyperSAGE with the transductive hypernetwork embedding method HGNN in HITTER to form the transductive version of HITTER, i.e., HITTER$_{trans}$. From Table \ref{tbl:overall-performacne} and Figure \ref{fig:detailed-anc-curve}, it can be found that HITTER$_{trans}$ is always among the worst methods. {Moreover,} from the comparison of the validation curves between HITTER and HITTER$_{trans}$ shown in Figure \ref{fig:validation-anc-curve}, we can find that HITTER$_{trans}$ does not converge {any more as a result of replacing the inductive hypernetwork embedding with a transductive one.}

{As known, an inductive hypernetwork embedding method can learn general topology patterns across different hypernetworks while a transductive one aims at learning specific topology patterns for a given hypernetwork. The different learning paradigms of these two kinds of hypernetwork embedding methods make them applicable for different kinds of tasks. Inductive hypernetwork embedding methods can be trained on some hypernetwork samples only once and generalized to embed the nodes of any unseen hypernetworks. While transductive hypernetwork embedding methods dig deeper into the topology of a specific hypernetwork and learns the embeddings of the nodes for this specific hypernetwork.\begin{figure}[htbp]
\centering
	\subfigtopskip=2pt
	\subfigure[HITTER]{
		\includegraphics[scale=0.25]{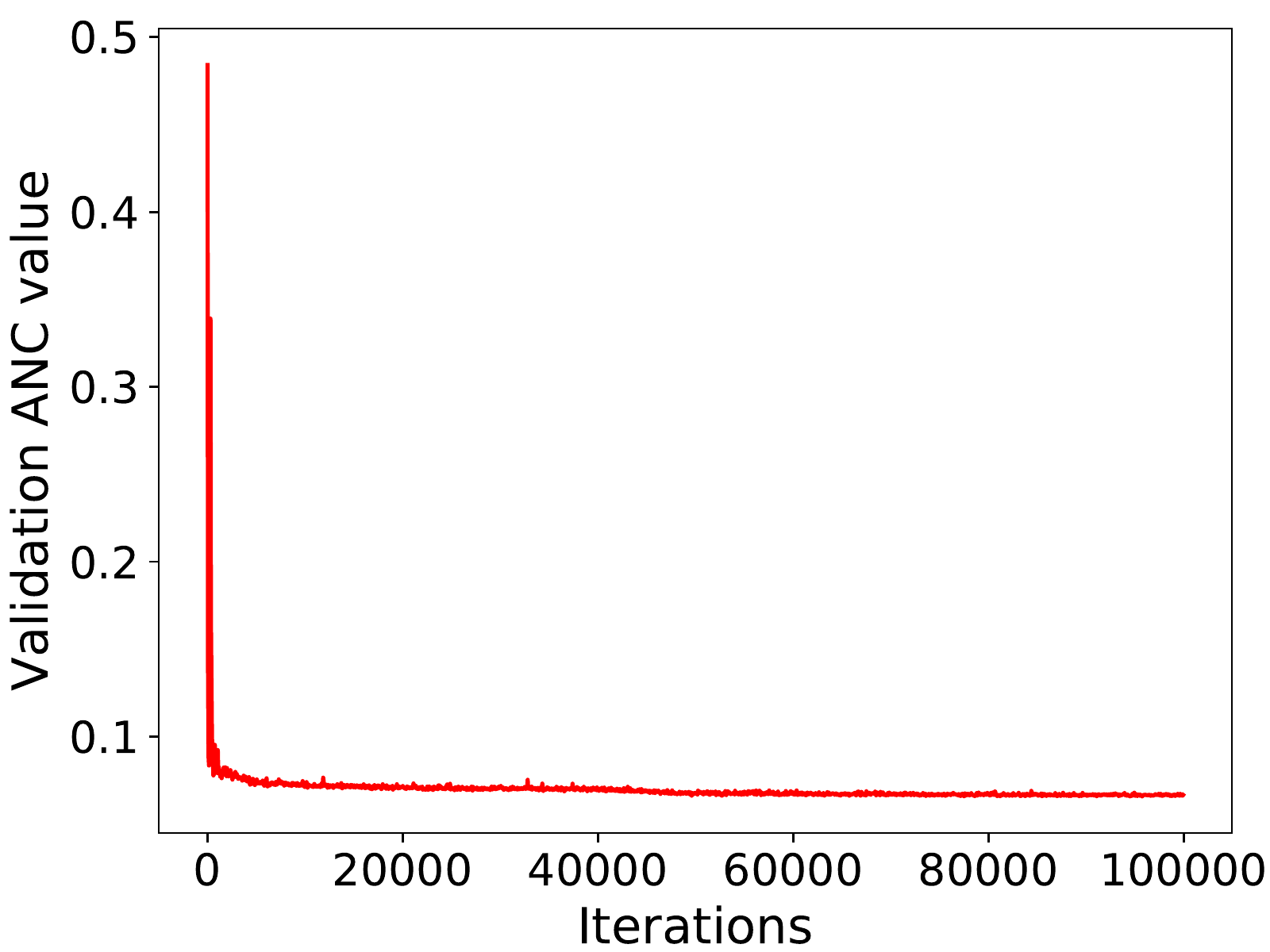}
		\label{5a}
	}
	\subfigure[HITTER$_{trans}$]{
		\includegraphics[scale=0.25]{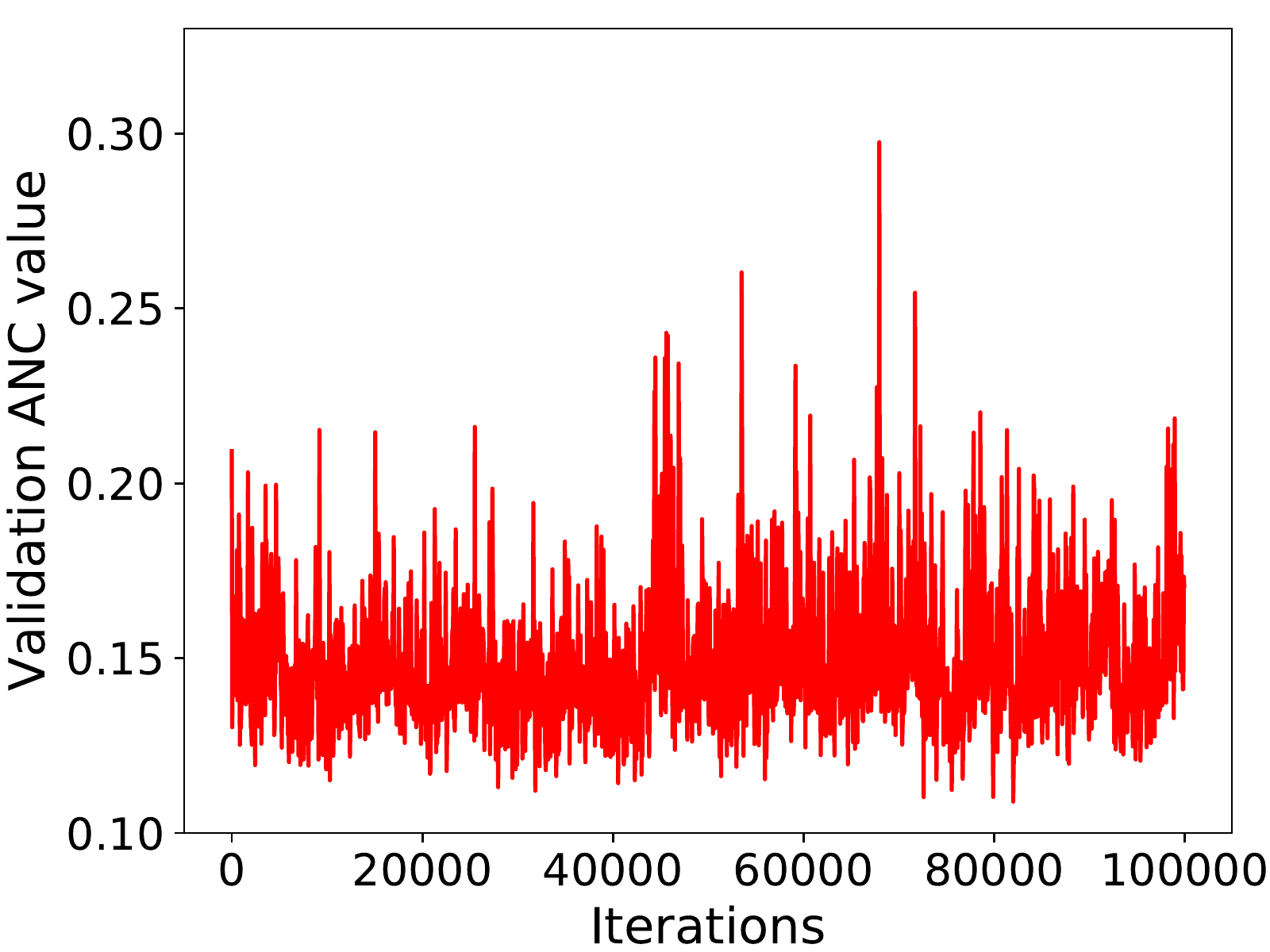}
		\label{5b}
	}
	\caption{Validation ANC curve of HITTER and HITTER$_{trans}$}
	\label{fig:validation-anc-curve}
\end{figure}
In our framework, HITTER is trained once on a large amount of synthetic hypernetworks and applied to dismantle unseen real-world hypernetworks. If transductive hypernetwork embedding method is used (i.e., HITTER$_{trans}$), for each episode in the training stage the newly generated hypernetwork will optimize the model parameters to fit its own specific topology, causing model parameters updated unsteadily across different episodes. Thus, HITTER$_{trans}$ does not converge as a result of transductive hypernetwork embedding and is always among the worst method.}

\subsection{Hyperparameter Analysis {(RQ3)}}

To ensure the flexibility, HITTER employs several hyperparameters for performance tuning. Among these hyperparameters, the embedding dimension and the balance of reconstruction loss are the most important ones. In this section, we conduct the analysis experiments on all the datasets used in this paper to reveal their effect on the performance {(i.e., RQ3)}.
\begin{figure*}[htbp]
\centering
	\subfigtopskip=2pt
	\subfigure[{Cora-co-authorship}]{
		\includegraphics[scale=0.35]{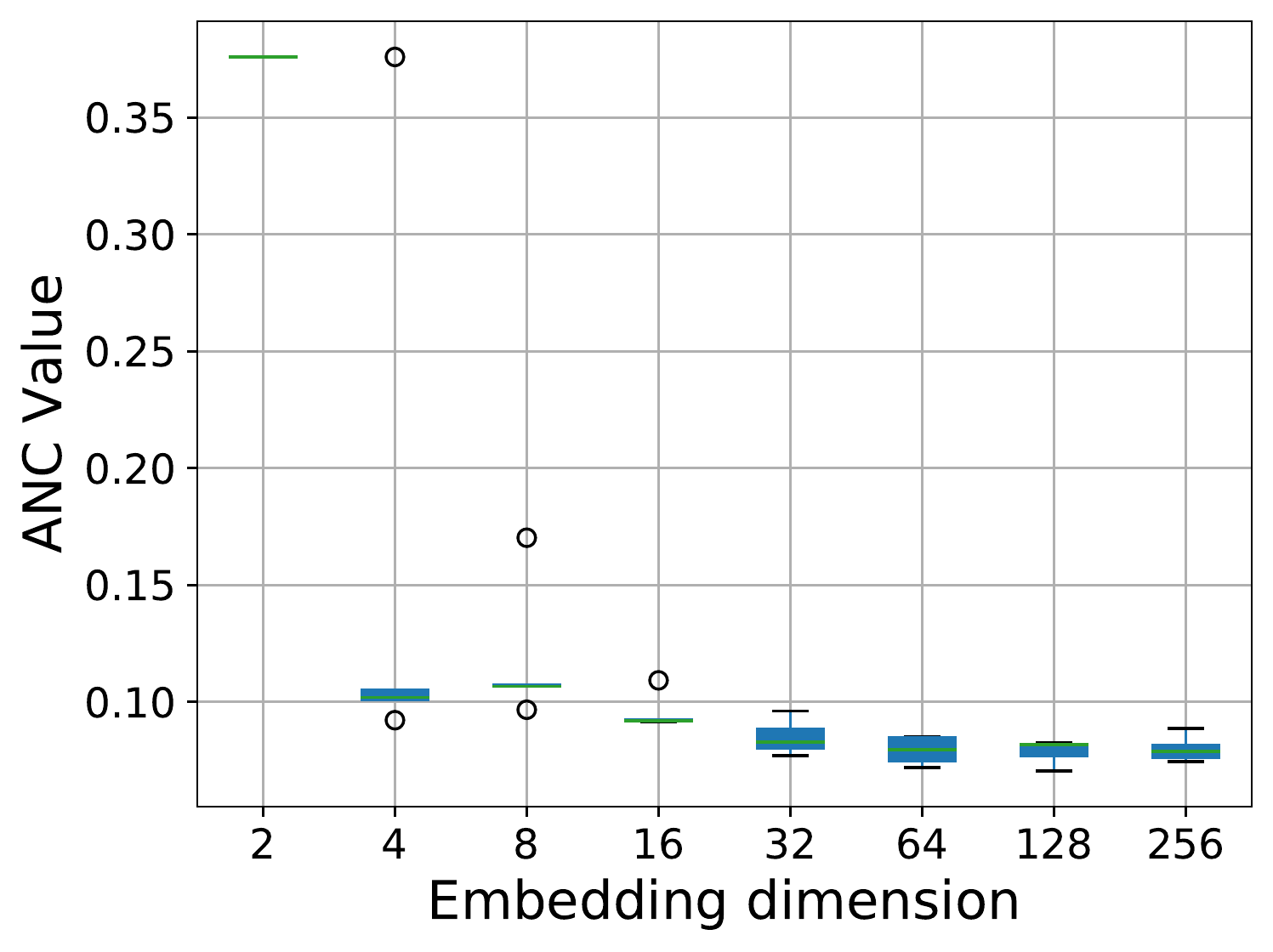}
		\label{fig:ed-cora}
	}
	\subfigure[MAG]{
		\includegraphics[scale=0.35]{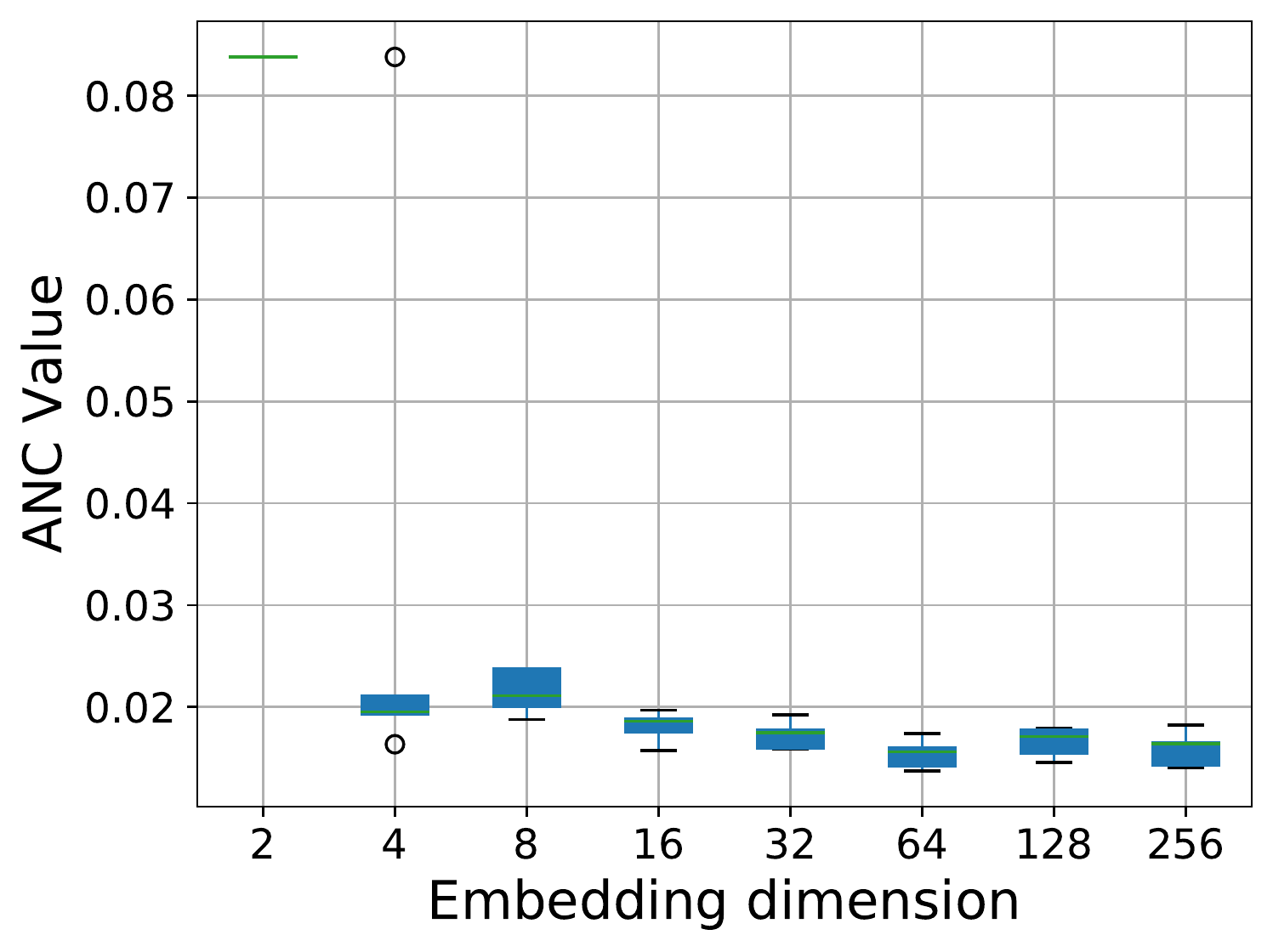}
		\label{fig:ed-mag}
	}
	\subfigure[Citeseer]{
		\includegraphics[scale=0.35]{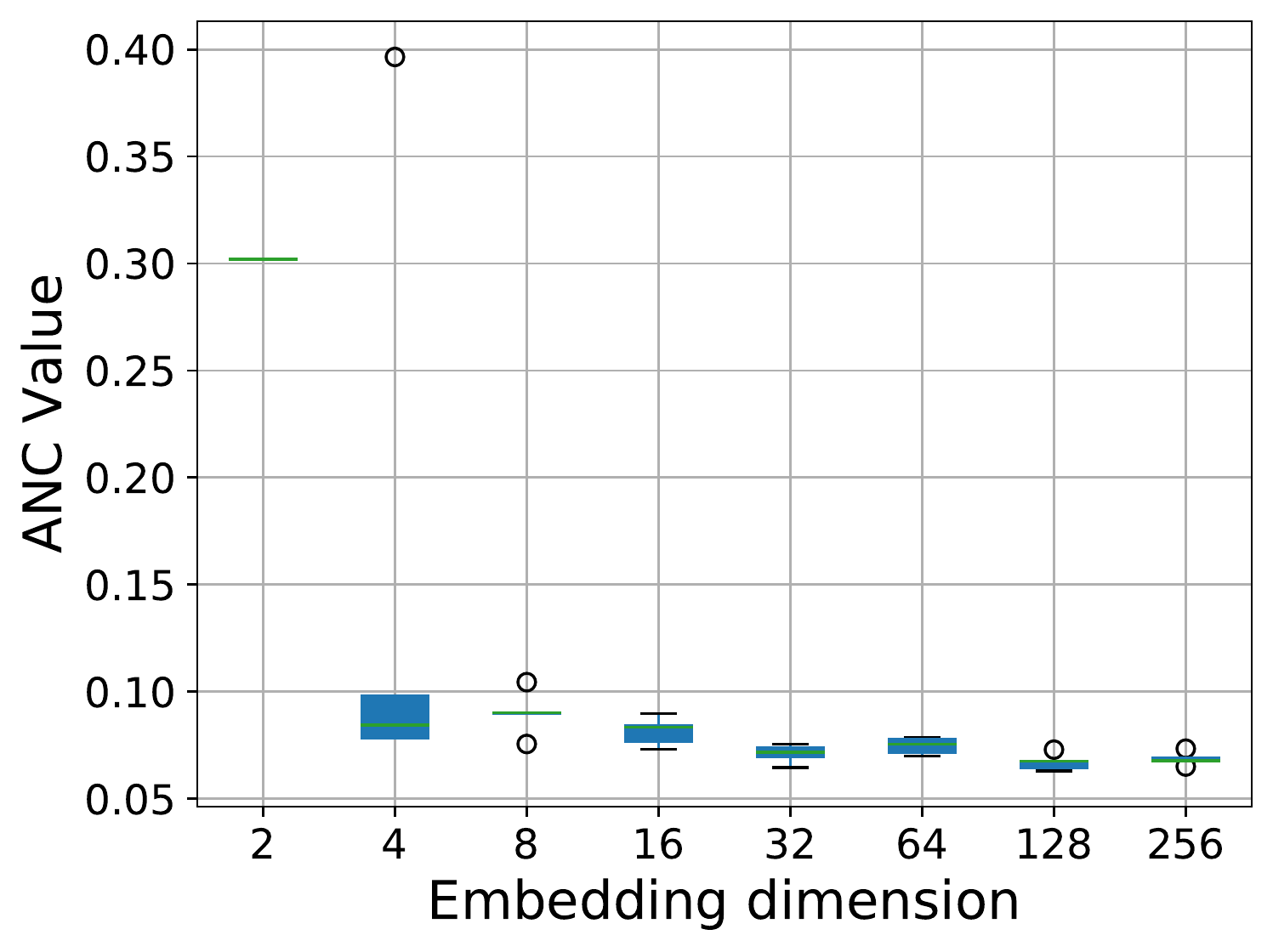}
		\label{fig:ed-citeseer}
	}
	\subfigure[Pubmed]{
		\includegraphics[scale=0.35]{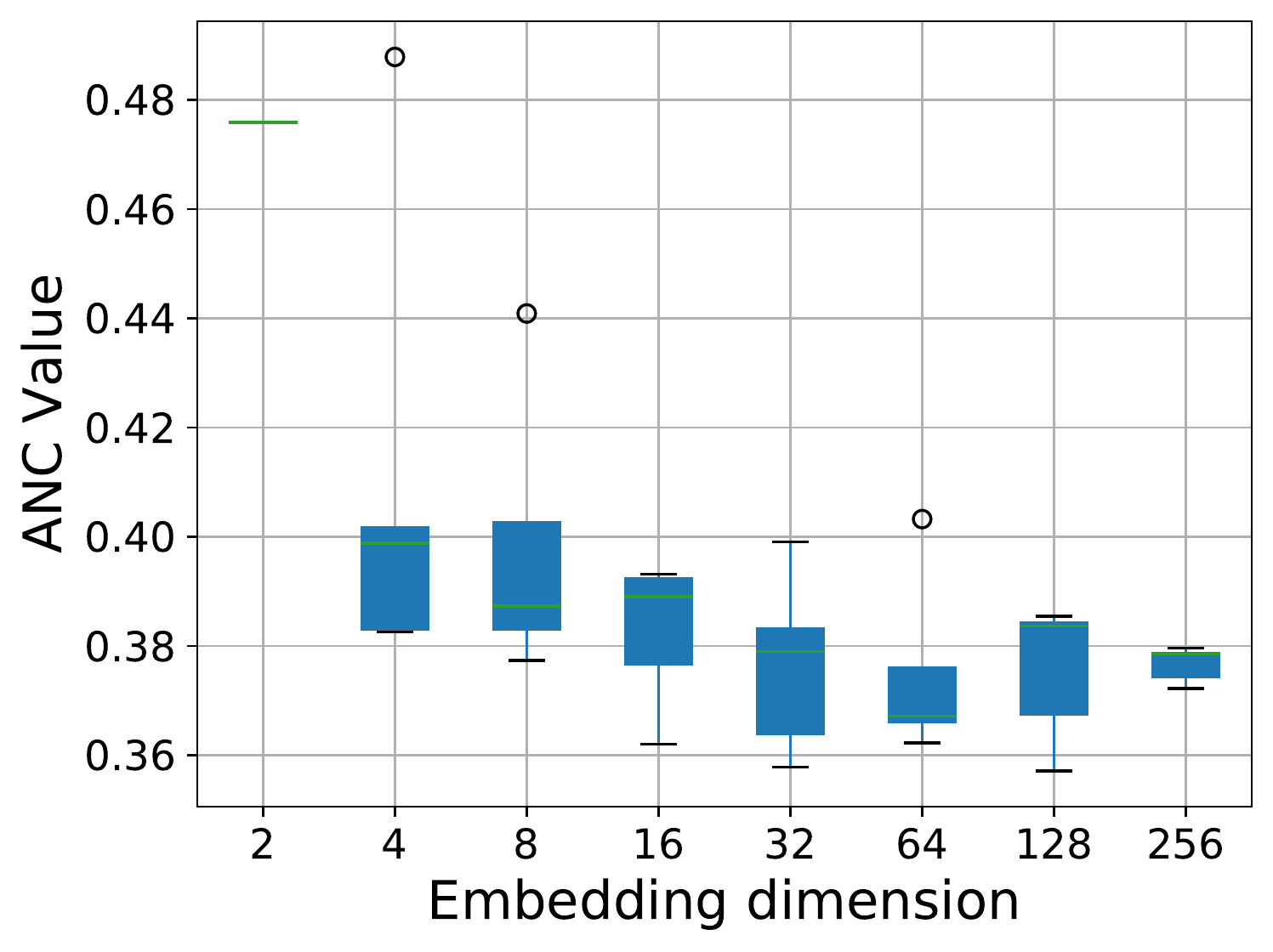}
		\label{fig:ed-pubmed}
	}
	\subfigure[NDC]{
		\includegraphics[scale=0.35]{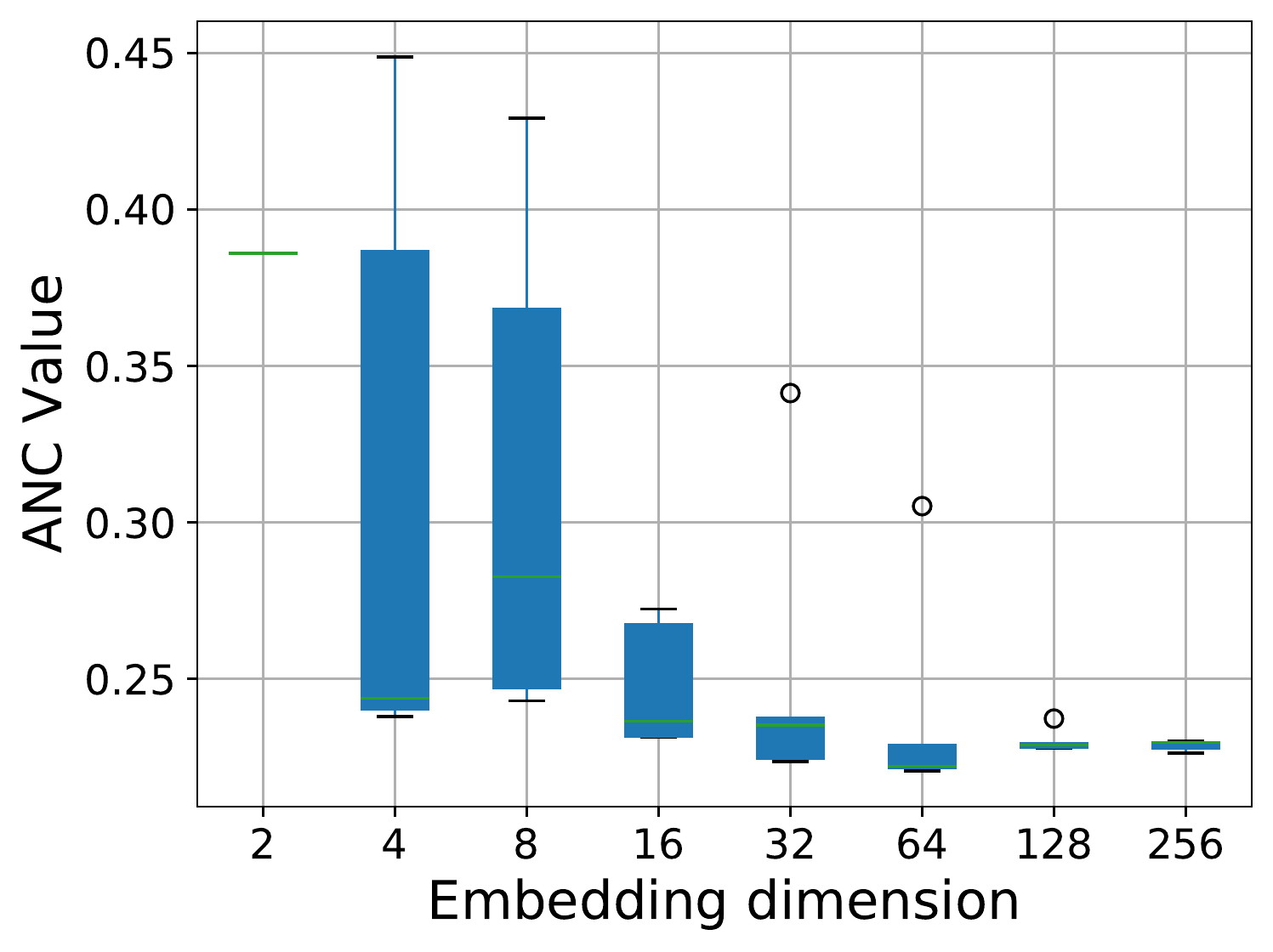}
		\label{fig:ed-ndc}
	}
	\caption{Impact of the embedding dimension}
	\label{fig:effect-dimension}
\end{figure*}
Firstly, embedding dimension usually determines the volume and density of information a model can capture and utilize. An unsuitable embedding dimension may decrease the performance or consume too much computation resources. In the analysis experiment, we train the model with different embedding dimensions ranging from 2 to 256 while keeping the other settings the same as those depicted in Section \ref{sec:experiment-dataset-settings}. To avoid the impact of randomness the experiments are repeated five times for each embedding dimension. Then the well-trained models are evaluated on all the five real-world datasets and the results of ANC values are summarized in Figure \ref{fig:effect-dimension}.

From the results shown in Figure \ref{fig:effect-dimension}, we can find that the average performance increases with the embedding dimension {in general}, which indicates that larger embedding dimension has a more powerful capacity to capture the complex structural information of the hypernetworks. The more structural information the embeddings have, the more guidance the DRL-based agent can learn from to obtain an optimal hypernetwork dismantling strategy. On the contrary, smaller embedding dimension smooths the structural difference of nodes, which makes it difficult for the DRL-based agent to distinguish the role of each node in maintaining the connectivity of the hypernetwork. Particularly the performance of the models with embedding dimension of 2 is extremely poor. However, it is not always efficient to have too large embedding dimension. As shown in Figure \ref{fig:effect-dimension}, {when the embedding dimension increases from 32 to 256 the growth of performance slows down and the performance on Pubmed and NDC even gets slightly worse. This is because a certain embedding dimension (e.g, 64 in most cases of our experiment) can capture enough topology information while more additional dimensions can only extract limited additional topology information or even introduce noise. Moreover, larger embedding dimension usually leads to higher computational cost.}

Moreover, in the view of robustness, we can also observe from Figure \ref{fig:effect-dimension} that the stability of performances is positively correlated with the embedding dimension in general. But the robustness of model performance on both Pubmed and NDC datasets are not as stable as that on {Cora-co-authorship}, MAG and Citeseer. {The main reason is that the hypernetworks of Pubmed and NDC are densely connected and the embeddings of many nodes learned by our proposed HyperSAGE become similar. Thus, during each step of the dismantling process of HITTER many nodes may be assigned to the same q-value by the DQN which uses the node embedding as input and a random one is selected for dismantling. This will result in many different sets of nodes output by HITTER as the optimal one, which increases the randomness and decreases the robustness of HITTER.}

\begin{figure*}[htbp]
\centering
	\subfigtopskip=2pt
	\subfigure[{Cora-co-authorship}]{
		\includegraphics[scale=0.35]{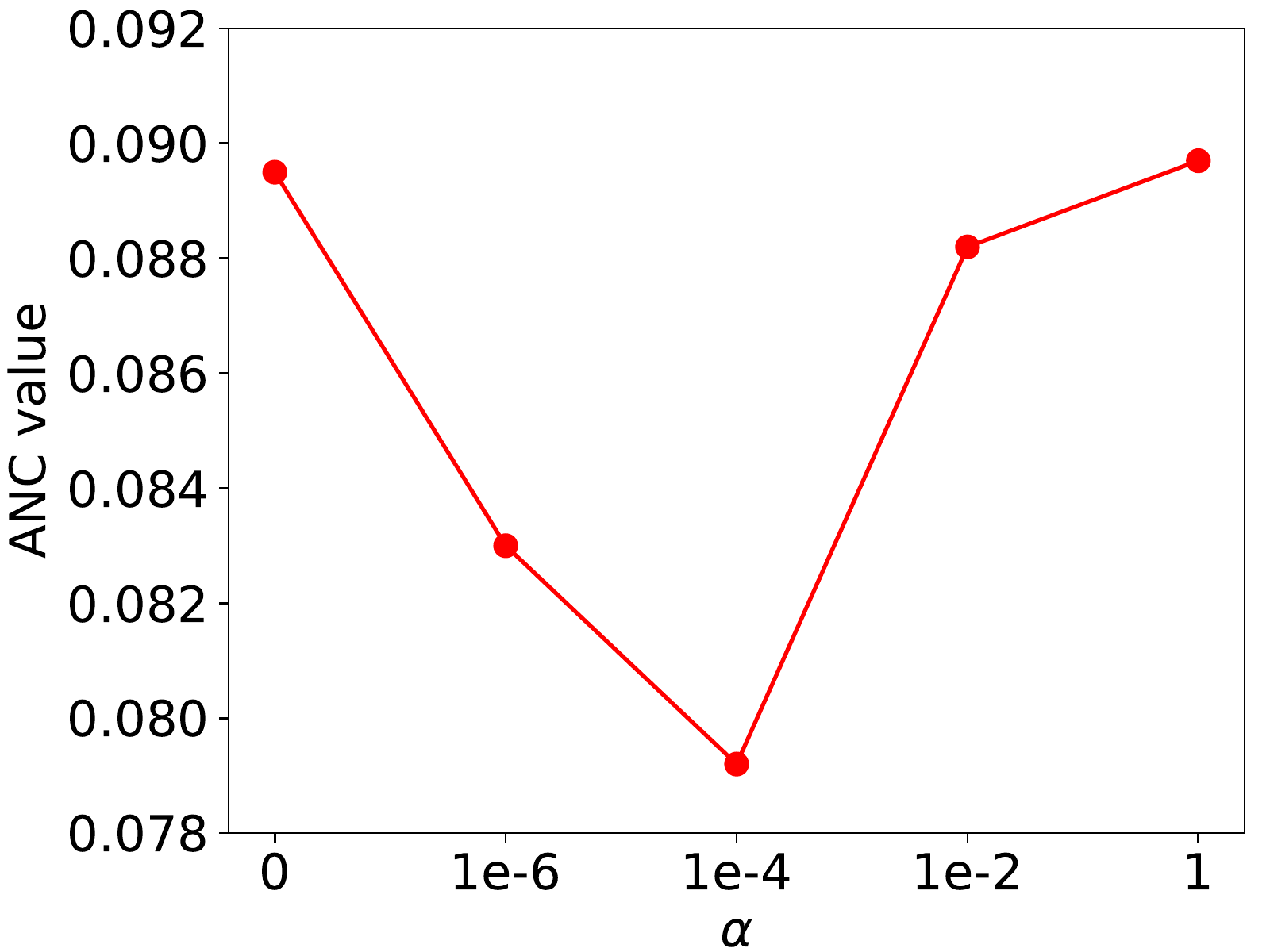}
	}
	\subfigure[MAG]{
		\includegraphics[scale=0.35]{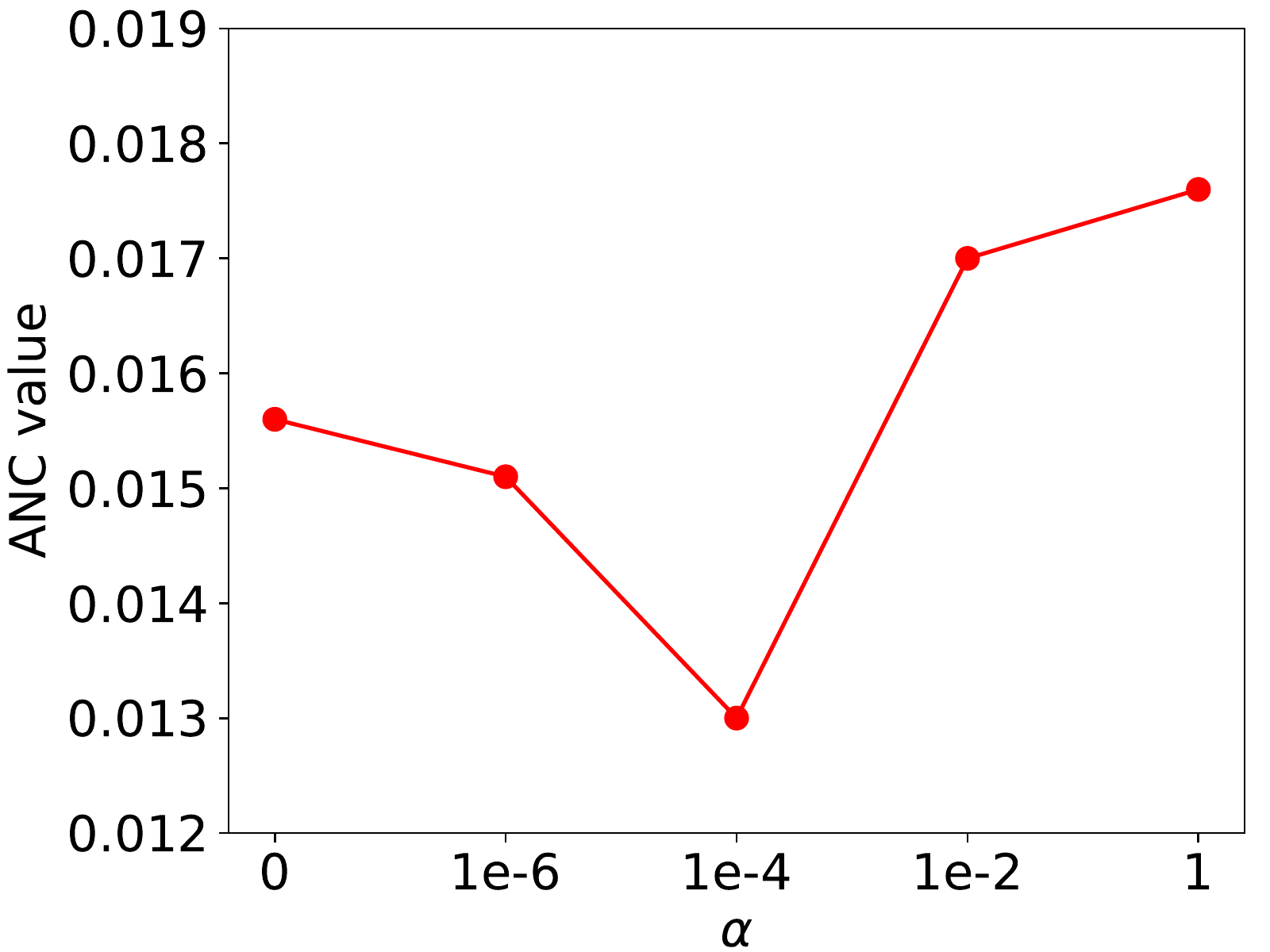}
	}
	\subfigure[Citeseer]{
		\includegraphics[scale=0.35]{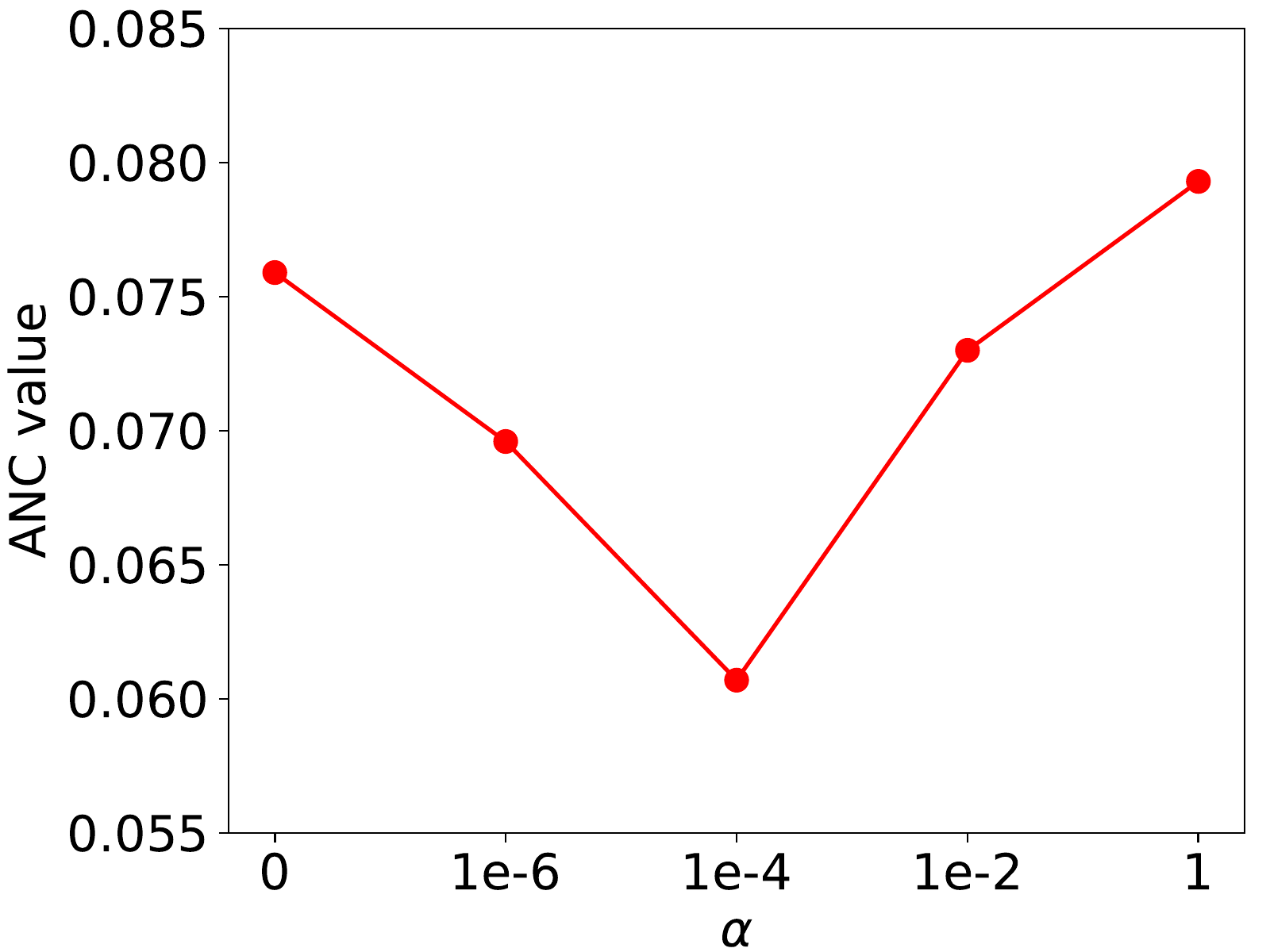}
	}
	\subfigure[Pubmed]{
		\includegraphics[scale=0.35]{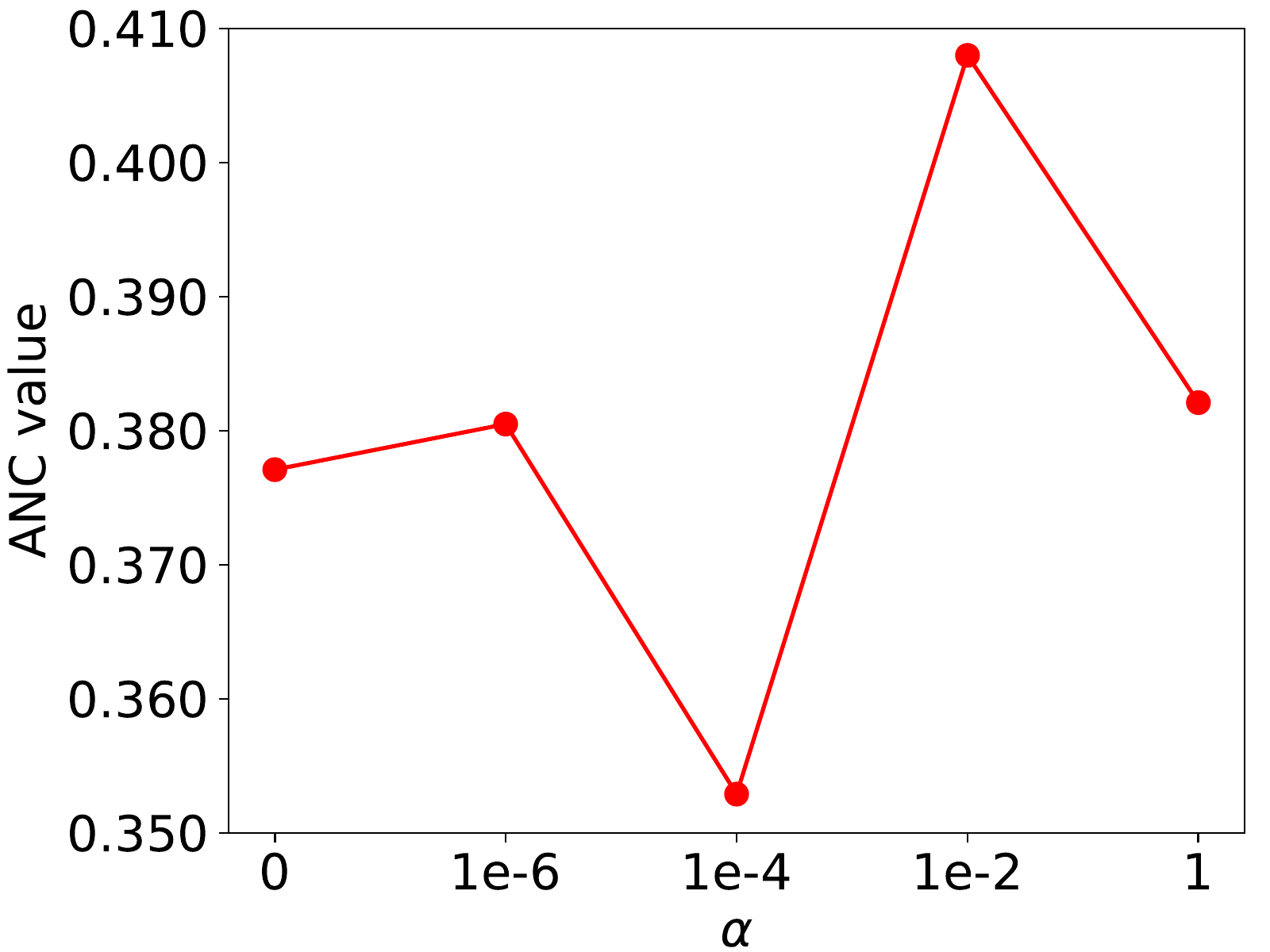}
	}
	\subfigure[NDC]{
		\includegraphics[scale=0.35]{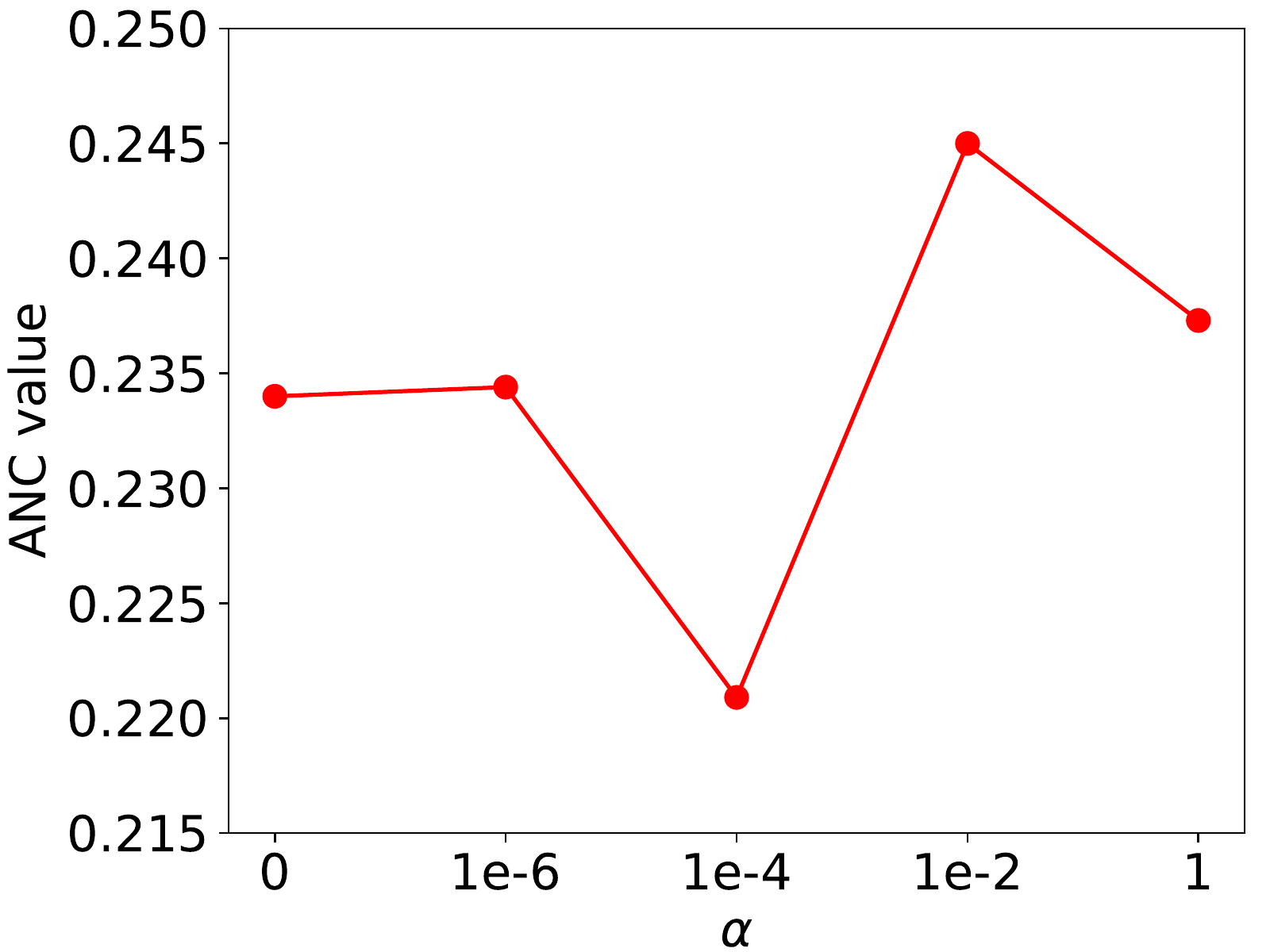}
	}
	\caption{Effect of the reconstruction loss of hypernetwork embedding}
	\label{fig:effect-alpha}
\end{figure*}
As another important hyperparamter, the ratio of reconstruction loss $\alpha$ balances the importance of hypernetwork embedding task and the hypernetwork dismantling task as HITTER learns them jointly. HITTER builds an end-to-end process which first learns the embeddings of the nodes and hypernetwork, and then feeds them to the DRL-based hypernetwork dismantling agent. The total loss of HITTER is composed of these two parts, i.e., the mean-square loss of deep Q network and the reconstruction loss of hypernetwork embedding. They jointly guide the learning of an optimal dismantling strategy, and different ratios of their contributions to the total loss represent different guidance.

\begin{figure*}[htbp]
\centering
	\subfigtopskip=2pt
	\subfigure[HITTER]{
		\includegraphics[scale=0.18]{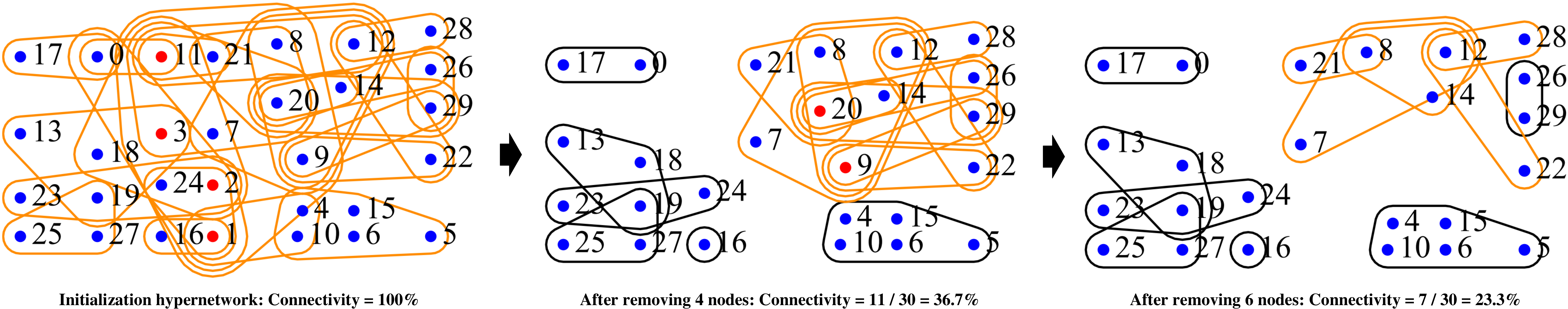}
		\label{example_hitter}
	}
	\subfigure[HHDA]{
		\includegraphics[scale=0.18]{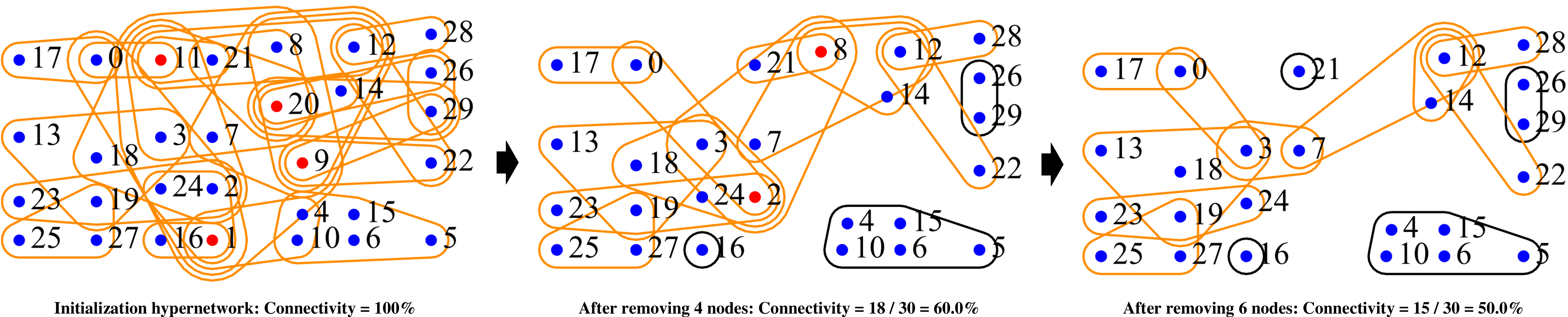}
		\label{example_hhda}
	}
	\caption{Visualization of dismantling steps of HITTER and HHDA. The hyperedges in orange belongs to the GCC in and nodes in red are selected to be removed.}
	\label{fig:vis}
\end{figure*}
In this analysis experiment, we train models with different ratios of reconstruction loss $\alpha$ and evaluate their performance on all the five real-world datasets. The results of ANC values are shown in Figure \ref{fig:effect-alpha}. The results demonstrate an obvious effect of the reconstruction loss on the performance of HITTER. {Too small and too large $\alpha$ both decrease the model performance}. {Usually}, a small $\alpha$ overlooks the importance of hypernetwork embedding and cannot capture enough structural information for the subsequent DRL-based hypernetwork dismantling agent while a large $\alpha$ decreases the significance of the agent and cannot train the deep Q network adequately. This is usually a universal phenomenon in various deep learning based models because a good performance of a model is determined by both good feature representations and adequate training of the model.
%{Moreover, the model performance in datasets Pubmed and NDC has an oscillation when $\alpha$ increase from 0.0001 to 1. We think it is mainly because the unstable performance of model on dense hypernetworks.}

\subsection{Visualization of Dismantling Steps}

To intuitively illustrate the advantages of our proposed HITTER, we visualize the dismantling steps of both HITTER and the second best performed method HHDA on a synthetic hypernetwork with 30 nodes. Initially all the 30 nodes belong to a single GCC and the connectivity of the hypernetwork is $100\%$. In each step, a batch of nodes are removed and the detailed removal processes are shown in Figure \ref{fig:vis}. 

Comparing the structures of the residual hypernetworks after each dismantling step of HITTER and HDDA, we can have the intuition that HITTER tends to destroy the connectivity sharply and considers less of the density of residual GCC while HHDA shows just the opposite. For example, in the first step HITTER chooses nodes $1$, $2$, $3$ and $11$ to remove while HHDA chooses nodes $1$, $9$, $11$ and $20$ to remove. Besides the common removal nodes $1$ and $11$, HITTER chooses nodes $2$ and $3$ instead of nodes $9$ and $20$ because nodes $2$ and $3$ are more responsible for connecting different parts of the hypernetwork while $9$ and $20$ plays more important role in forming a dense local part. Thus the connectivity of the residual hypernetwork dismantled by HITTER reduces sharply to $36.7\%$ while that dismantled by HHDA reduces smoothly to $60.0 \%$ in the first step. This also provides an intuitive explanation why the ANC curves of HITTER decreases more sharply in Figure \ref{fig:detailed-anc-curve}.

It is also interesting to point out that from a global perspective only one node is different between the removed node sets of HITTER and HHDA. The symmetric difference of these two node sets is $\{3, 8\}$. But the final connectivity of the residual hypernetworks is significantly different. This indicates that in a budget limited dismantling process HITTER is more effective than HHDA.

\subsection{{Case Study (RQ4)}}

{Hypernetwork dismantling has many practical applications, one of which is epidemic containment. As epidemics like COVID-19 usually outbreak through human physical contact network, it is an effective epidemic containment strategy to fragment the network into many small unconnected pieces by immunizing an optimal set of persons who play important role in maintaining the connectivity of the network. And network dismantling is an effective method to find this optimal set of persons. However, conventional network dismantling methods take no consideration of the effect of group contact on epidemic outbreaks while this kind of cases widely exist. For example, a recent COVID-19 outbreak is originated from the group contacts of mahjong players in a large room in Yangzhou, China.}

{Our proposed hypernetwork dismantling model HITTER can effectively deal with this practical situation by modeling group contact as hypernetwork. To demonstrate the effectiveness of our model, we conduct the classical SIR simulation on a real-world contact dataset \cite{listcontact} from the Science Gallery in Dublin, Ireland. A hypernetwork is first constructed from this dataset with a group of persons who visited the Science Gallery at the same timestamp as a hyperedge. Then for each of the hypernetwork dismantling methods including HITTER, an optimal set of persons is selected and marked as immunized persons who 
\begin{table}
\caption{{Final infection rate after hypernetwork dismantling.}}
\label{caseStudy}
\centering
\renewcommand\arraystretch{1.5}
\setlength{\tabcolsep}{2.0mm}
\begin{tabular}{cccccc}
\toprule
Immune Ratio & 0\%     & 5\%     & 10\%    & 15\%    & 20\%    \\ \toprule
HD          & 23.62\% & 10.99\% & 6.66\%  & 4.82\%  & 4.44\%  \\
CI          & 23.62\% & 13.30\% & 9.61\%  & 6.57\%  & 4.39\%  \\
GND         & 23.62\% & 12.62\% & 13.30\% & 12.44\% & 10.55\% \\
FINDER      & 23.62\% & 21.62\% & 20.20\% & 20.81\% & 18.83\% \\
HHD         & 23.62\% & 9.28\%  & 6.99\%  & 4.94\%  & 3.75\%  \\
SubTSSH     & 23.62\% & 10.90\% & 6.13\%  & 6.03\%  & 5.24\%  \\
HITTER      & 23.62\% & 7.68\%  & 4.97\%  & 3.78\%  & 3.39\%  \\ \toprule
\end{tabular}
\end{table}
will not be infected in the SIR simulation. Finally, an SIR simulation is conducted with the persons in the earliest contact group as infect. Due to the infectiousness and recover rate vary widely between different infection diseases, and this simulation experiment concentrates on the influence of hypernetwork dismantling on epidemic control, not special disease, the infection and recover probability are both set to 0.1. For each hypernetwork dismantling methods, different portions of immunized persons are tested, ranging from 0 to 0.2. To avoid the randomness of the SIR simulation, each simulation is repeated 100 times and the average final infection rate is calculated. The results are shown in Table \ref{caseStudy}.}

The results of SIR simulation demonstrate the effectiveness of hypernetwork dismantling on epidemic containment. With the increase of the portion of immunized persons, the final infection rate can be decreased. Most of these methods can reduce the final infection rate to a very low level through immunizing just 10 percent of persons. Moreover, among these dismantling methods our proposed HITTER has the best performance as it effectively utilizes the group contact information by hypernetwork model and learns better dismantling strategy with the power of deep reinforcement learning.

\section{Conclusion and Future Work} \label{sec:conclusion}

In this paper, we study the hypernetwork dismantling problem. Specifically, considering the NP-hard nature of this problem we formulated it from a deep reinforcement learning perspective and proposed a novel framework HITTER by incorporating the inductive hypernetwork embedding and deep Q learning techniques. Comprehensive experiments were conducted on five real-world datasets from diverse domains. Our experimental results demonstrated the effectiveness of our proposed framework compared with baselines of either simple networks or greedy paradigms for hypernetworks. Moreover, the necessity of the inductive hypernetwork embedding method for good transferability is also validated.

{In future work, we intend to improve the robustness of the model on densely connected hypernetworks and propose robust hypernetwork learning method or hypernetwork sparsity method to learn more robust embeddings which discriminate nodes in dense hypernetwork more clearly.}
\bibliographystyle{IEEEtran}
\bibliography{HITTER}


\begin{IEEEbiography}[{\includegraphics[width=1in,height=1.25in,clip,keepaspectratio]{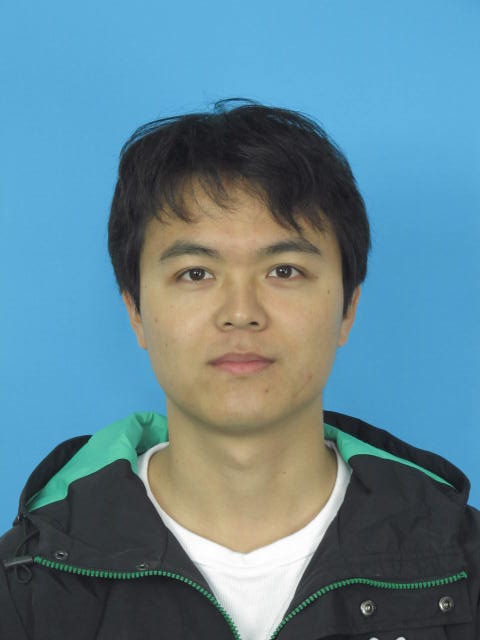}}]{Dengcheng Yan}
received the B.S. and Ph.D. degrees from the University of Science and Technology of China, in 2011 and 2017, respectively. From 2017 to 2018, he was a Core Technology Researcher and a Big Data Engineer with Research Institute of Big Data, iFlytek Company Ltd. He is currently a Lecturer with Anhui University, China. His research interests include software engineering, recommendation systems and complex networks.
\end{IEEEbiography}

\begin{IEEEbiography}[{\includegraphics[width=1in,height=1.25in,clip,keepaspectratio]{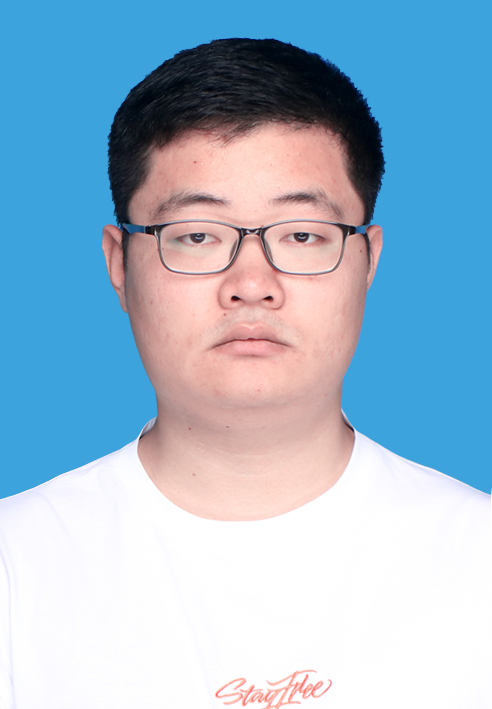}}]{Wenxin Xie}
received the B. S. degree from Liaoning Petrochemical University, China, in 2018. He is currently pursuing the M. S. degree with Anhui University of computer science and technology. His research interest includes graph combinatorial optimization.
\end{IEEEbiography}

%\newpage

\begin{IEEEbiography}[{\includegraphics[width=1in,height=1.25in,clip,keepaspectratio]{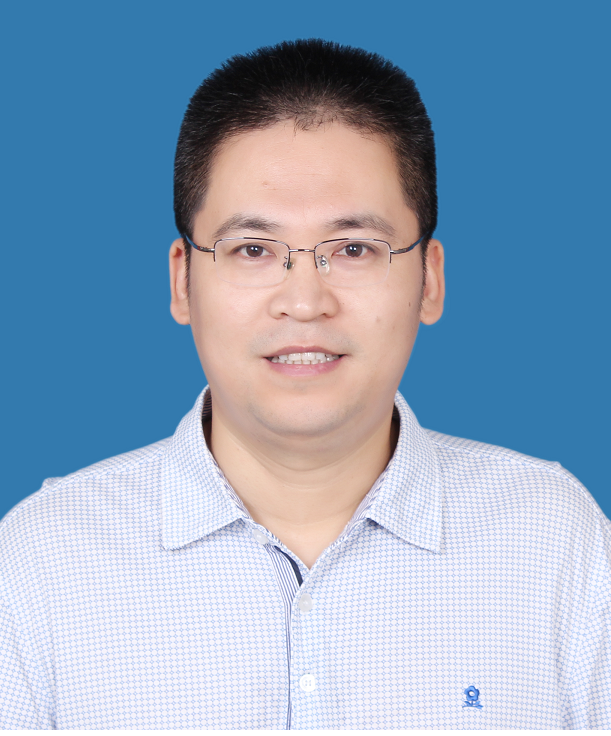}}]{Yiwen Zhang}
received the Ph.D. degree in management science and engineering from the Hefei University of Technology, in 2013. He is currently a Professor with the School of Computer Science and Technology, Anhui University. His research interests include service computing, cloud computing, and big data analytics.
\end{IEEEbiography}

\begin{IEEEbiography}[{\includegraphics[width=1in,height=1.25in,clip,keepaspectratio]{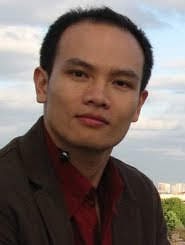}}]{Qiang He}
received his first PhD degree from Swinburne University of Technology, Australia, in 2009 and his second PhD degree in computer science and engineering from Huazhong University of Science and Technology, China, in 2010. He is a senior lecturer at Swinburne. His research interests include service computing, software engineering, cloud computing and edge computing.
\end{IEEEbiography}

\begin{IEEEbiography}[{\includegraphics[width=1in,height=1.25in,clip,keepaspectratio]{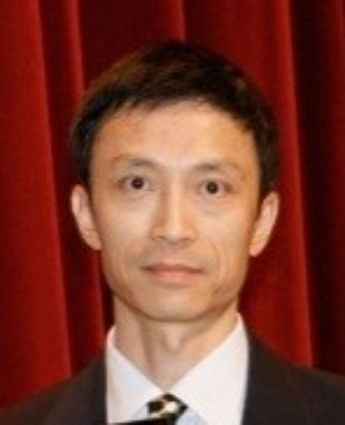}}]{Yun Yang} received his PhD degree from the
University of Queensland, Australia, in 1992, in computer science. He is currently a full professor in the School of Software and Electrical Engineering at Swinburne University of Technology, Melbourne, Australia. His research interests include software technologies, cloud and edge computing, workflow systems, and service computing.
\end{IEEEbiography}


%\vfill


\end{document}